\documentclass[10pt,twocolumn,letterpaper]{article}

\usepackage{pdfpages}
\usepackage{multido}

\usepackage{iccv}
\usepackage{times}
\usepackage{epsfig}
\usepackage{graphicx}
\usepackage{amsmath}
\usepackage{amssymb}
\usepackage{overpic}
\usepackage{booktabs, multirow} 
\usepackage{changepage,threeparttable} 
\usepackage{adjustbox}
\usepackage[table,xcdraw,dvipsnames]{xcolor}

\usepackage{caption}
\usepackage{comment}

\newcommand{\mc}[3]{\multicolumn{#1}{#2}{#3}}

\usepackage[breaklinks=true,letterpaper=true,colorlinks,bookmarks=false]{hyperref}

\newcommand{\notsosmall}{\fontsize{10.5pt}{12pt}\selectfont}

\newcommand\blfootnote[1]{%
  \begingroup
  \renewcommand\thefootnote{}\footnote{#1}%
  \addtocounter{footnote}{-1}%
  \endgroup
}

\usepackage[capitalize]{cleveref}
\crefname{section}{Sec.}{Secs.}
\Crefname{section}{Section}{Sections}
\Crefname{table}{Table}{Tables}
\crefname{table}{Tab.}{Tabs.}

\iccvfinalcopy

\definecolor{somegray}{rgb}{0.5, 0.5, 0.5}
\newcommand{\darkgrayed}[1]{\textcolor{somegray}{#1}}
\makeatletter
\newcommand*\titleheader[1]{\gdef\@titleheader{#1}}
\AtBeginDocument{%
  \let\st@red@title\@title
  \def\@title{%
    \vskip-3em
    \bgroup\normalfont\large\centering\@titleheader\par\egroup
    \vskip1.5em\st@red@title}
}
\makeatother

\titleheader{\darkgrayed{This paper has been accepted for publication at the \\
IEEE/CVF International Conference on Computer Vision (ICCV), Paris, 2023.
\copyright IEEE}}

\title{\includegraphics[scale=0.04]{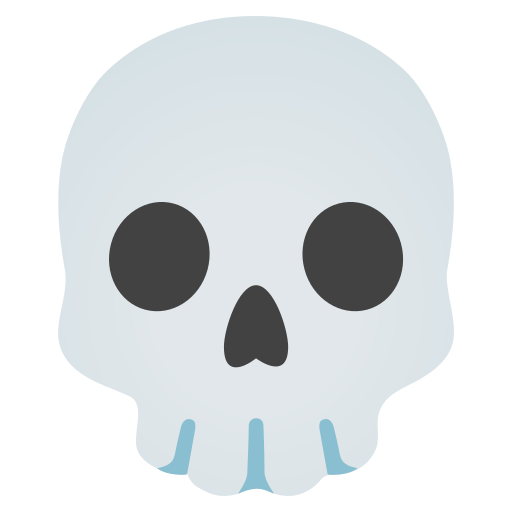} 
To Adapt or Not to Adapt? \\ Real-Time Adaptation for Semantic Segmentation}

\begin{document}

\author{Marc Botet Colomer$^*$ $^{1,2}$  \hspace{1cm} Pier Luigi Dovesi$^*$ $^3$ $^\dagger$ \\  Theodoros Panagiotakopoulos$^4$ \hspace{1cm} Joao Frederico Carvalho$^1$ \hspace{1cm} Linus Härenstam-Nielsen$^{5,6}$ \\ Hossein Azizpour$^2$ \hspace{1cm} Hedvig Kjellström$^{2,3}$ \hspace{1cm} Daniel Cremers$^{5,6,7}$ \hspace{1cm} Matteo Poggi$^8$ \vspace{0.3cm}\\ 
\notsosmall $^1$Univrses \hspace{1cm} $^2$KTH \hspace{1cm} $^3$Silo AI  \hspace{1cm} $^4$King  \hspace{1cm} $^5$Technical University of Munich \\ \notsosmall $^6$Munich Center for Machine Learning \hspace{1cm} 
$^7$University of Oxford \hspace{1cm} $^8$University of Bologna \\
\small{\url{https://marcbotet.github.io/hamlet-web/}}
}

\setcounter{figure}{1}

\twocolumn[{
\renewcommand\twocolumn[1][]{#1}
\maketitle
\begin{center}
    \vspace{-0.5cm}
    \includegraphics[trim=0cm 0.2cm 0cm 0cm, clip,width=\textwidth]{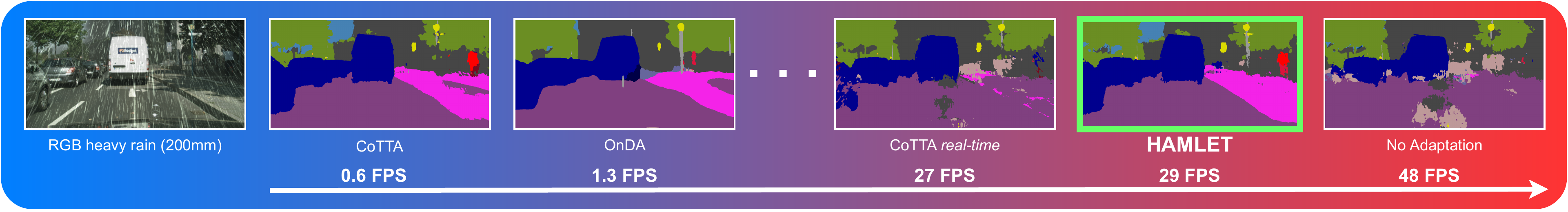} 
    \label{fig:teaser}
\end{center}
\vspace{-0.4cm}

\small \hypertarget{fig:teaser}{Figure 1.} \textbf{Real-time adaptation with HAMLET.} Online adaptation to continuous and unforeseeable domain shifts is hard and computationally expensive. HAMLET can deal with it at almost 30FPS outperforming much slower online methods -- \textit{\eg{}} OnDA and CoTTA.
\vspace{0.2cm}
}]

\maketitle

\begin{abstract}
   The goal of Online Domain Adaptation for semantic segmentation is to handle unforeseeable domain changes that occur during deployment, like sudden weather events. However, the high computational costs associated with brute-force adaptation make this paradigm unfeasible for real-world applications.
   In this paper we propose HAMLET, a Hardware-Aware Modular Least Expensive Training framework for real-time domain adaptation. Our approach includes a hardware-aware back-propagation orchestration agent (HAMT) and a dedicated domain-shift detector that enables active control over when and how the model is adapted (LT).
   Thanks to these advancements, our approach is capable of performing semantic segmentation while simultaneously adapting at more than 29FPS on a single consumer-grade GPU. Our framework's encouraging accuracy and speed trade-off is demonstrated on OnDA and SHIFT benchmarks through experimental results.
\end{abstract}

\section{Introduction}
\label{sec:intro}

Semantic segmentation aims at classifying an image at a pixel level, based on the local and global context, to enable a higher level of understanding of the depicted scene.\blfootnote{$^*$ Joint first authorship \hspace{0.3cm} $^\dagger$ Part of the work done while at Univrses}
In recent years, deep learning has become the dominant paradigm to tackle this task effectively employing CNNs~\cite{chen2017deeplab,yuan2021segmentation,chen2020naive} or, more recently, transformers~\cite{xie2021segformer}, at the expense of requiring large quantities of annotated images for training.
Specifically, annotating for this task needs per-pixel labeling, which is an expensive and time-consuming task, severely limiting the availability of training data.

\begin{figure}
    \centering
    \includegraphics[trim=0cm 1cm 0cm 1cm, clip, width=0.45\textwidth]{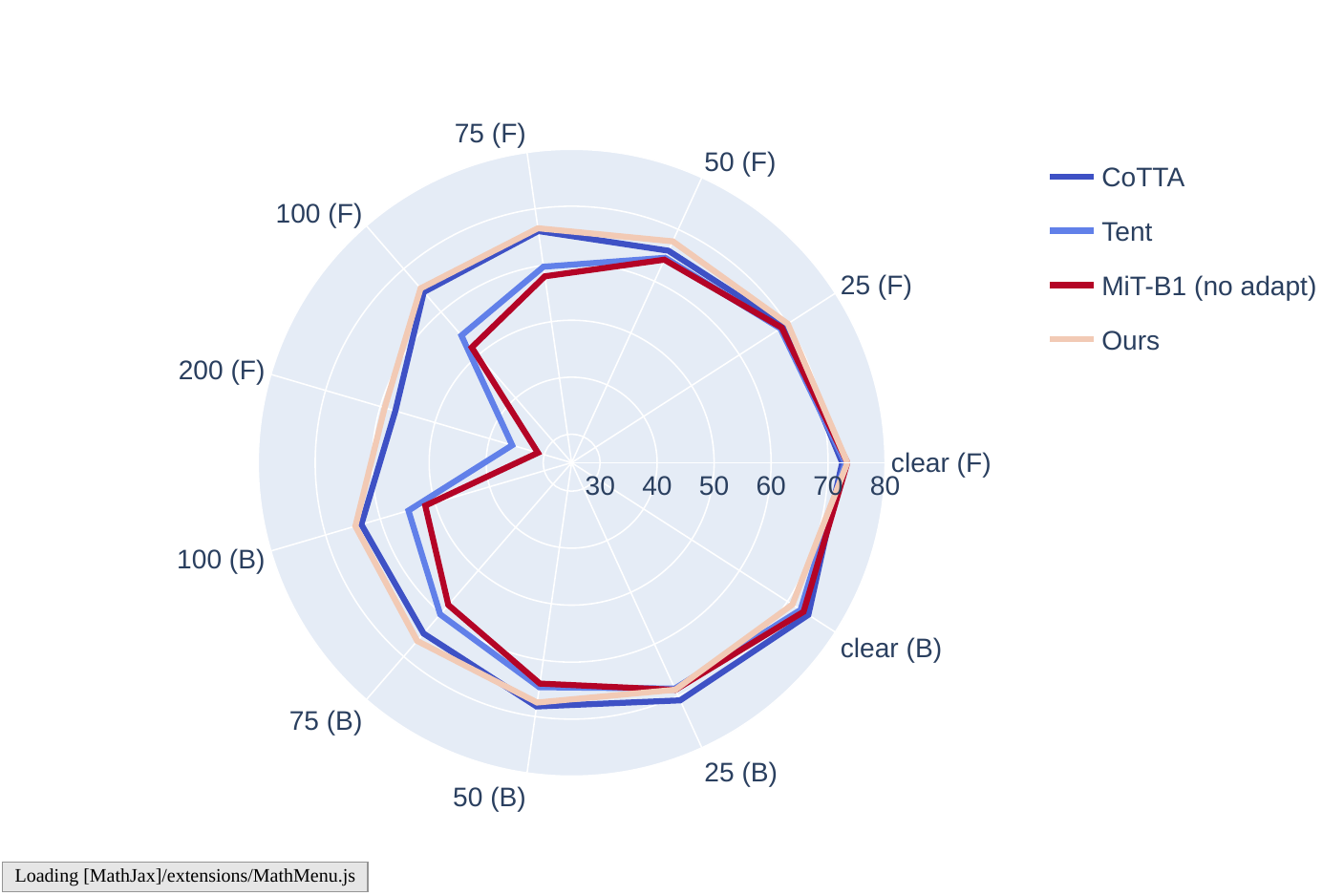}
    \vspace{-0.3cm}
    \caption{\textbf{Online adaptation methods on the Increasing Storm.} We plot mIoUs achieved on single domains. Colors from colder to warmer encode slower to faster methods.}
    \label{fig:radar}
\end{figure}

The use of simulations and graphics engines~\cite{gta} to generate annotated frames enabled a marked decrease in the time and cost necessary to gather labeled data thanks to the availability of the ground truth.
However, despite the increasing quality in data realism~\cite{thomas_pandikow_kim_stanley_grieve_2021}, there is a substantial difference between simulated data generated by graphics engines and real-world images, such that leveraging these data for real-world applications requires adapting over a significant domain shift.
The promise of unlocking this cheap and plentiful source of training data has provided a major impulse behind the development of a large body of work on Unsupervised Domain Adaptation (UDA) techniques~\cite{cyclegan,dcan,cycada,ganin,fada}, consisting of training semantic segmentation networks on labelled synthetic frames -- the \textit{source} domain -- and then adapting the network to operate on real images, representing the \textit{target} domain, without requiring human annotation.
However, the synthetic-to-real shift represents only one of many possible domain transitions; specifically, when dealing with real-world deployment, domain shifts can occur from various causes, from different camera placements to different lighting, weather conditions, urban scenario, or any possible combination of the above.
Because of the combinatorial nature of the problem, it is simply impossible to evenly represent all possible deployment domains in a dataset. This \textit{curse of dimensionality} prevents having generalized robust perfomances~\cite{Panagiotakopoulos_ECCV_2022,shift}.
However, the recent advent of \textit{online} domain adaptation~\cite{Panagiotakopoulos_ECCV_2022} potentially allows us to face continuous and unpredictable domain shifts at deployment time, without requiring data associated with such domain shifts beforehand.
Nonetheless, despite its potential, several severe limitations still hamper the online adaptation paradigm.
In particular, continuously performing back-propagation on a frame-by-frame schedule~\cite{Panagiotakopoulos_ECCV_2022} incurs a high computational cost, which negatively affects the performance of the network, dropping its overall framerate to accommodate the need for continuous adaptation.
Various factors are involved in this matter: first, the severity of this overhead is proportional to the complexity of the network itself -- the larger the number of parameters, the heavier the adaptation process becomes; second, we argue that frame-by-frame optimization is an excessive process for the adaptation itself -- not only the network might need much fewer optimization steps to effectively counter domain shifts, but also such an intense adaptation definitely increases the likelihood of catastrophic forgetting over previous domains~\cite{kirkpatrick_overcomming_2016, shift}.
In summary, a practical solution for online domain adaptation in semantic segmentation that can effectively operate in real-world environments and applications still seems to be a distant goal.

In this paper, we propose a novel framework aimed at overcoming these issues and thus allowing for real-time, online domain adaptation: 
\begin{itemize}
    \item We address the problem of online training by designing an automatic lightweight mechanism capable of significantly reducing back-propagation complexity. We exploit the model modularity to automatically choose to train the network subset which yields the highest improvement for the allocated optimisation time. This approach reduces back-propagation FLOPS by 34\% while minimizing the impact on accuracy.
    \item In an orthogonal fashion to the previous contribution, we introduce a lightweight domain detector. This allows us to design principled strategies to activate training only when it really matters as well as setting hyperparameters to maximize adaptation speed. Overall, these strategies increase our speed by over $5\times$ while sacrificing less than 2.6\% in mIoU.
    \item We evaluate our method on multiple online domain adaptation benchmarks both fully synthetic~\cite{shift} and semi-synthetic CityScapes domain sequences~\cite{Panagiotakopoulos_ECCV_2022}, showing superior accuracy and speed compared to other test-time adaptation strategies.
\end{itemize}

Fig. \hyperref[fig:teaser]{1} demonstrates the superior real-time adaptation performance of HAMLET compared to slower methods such as CoTTA~\cite{wang2022continual}, which experience significant drops in performance when forced to maintain a similar framerate by adapting only once every 50 frames. In contrast, HAMLET achieves an impressive 29 FPS while maintaining high accuracy. Additionally, Fig. \ref{fig:radar} offers a glimpse of HAMLET's performance on the Increasing Storm benchmark~\cite{Panagiotakopoulos_ECCV_2022}, further highlighting its favorable accuracy-speed trade-off.

\section{Related Work}

We review the literature relevant to our work, about semantic segmentation and UDA, with particular attention to continuous and online methodologies.

\textbf{Semantic Segmentation.} Very much like classification, deep learning plays a fundamental role in semantic segmentation. Fully Convolutional Network (FCN)~\cite{fcn} represents the pivotal step in this field, adapting common networks by means of learned upsample operators (deconvolutions).
Several works aimed at improving FCN both in terms of speed~\cite{yu2018bisenet,nekrasovlight} and accuracy~\cite{chen2017deeplab,deeplabv2,chen2018encoder}, with a large body of literature focusing on the latter. Major improvements have been achieved by enlarging the receptive field~\cite{zhao2017pspnet,yang2018denseaspp,chen2017deeplab,deeplabv2,chen2018encoder}, introducing refinement modules~\cite{fu2019adaptive,zhou2019context,he2019adaptive}, exploiting boundary cues~\cite{chen2016semantic,ding2019boundary,takikawa2019gated} or using attention mechanisms in different flavors~\cite{fu2019dual,li2019expectation,wang2018non,xie2021segmenting}. 
The recent spread of Transformers in computer vision~\cite{dosovitskiy2020image} reached semantic segmentation as well~\cite{xie2021segmenting,yuan2021segmentation,xie2021segformer}, with SegFormer~\cite{xie2021segformer} representing the state-of-the-art in the field and being the object of studies in the domain adaptation literature as well~\cite{hoyer2021daformer}.

\textbf{Unsupervised Domain Adaptation (UDA).}
This body of research aims at adapting a network trained on a \textit{source}, labeled domain to a \textit{target}, unlabeled one. Early approaches rely on the notion of ``style'' and learn how to transfer it across domains~\cite{cyclegan,dcan,cycada,bdl,stylization,yang_fda_2020}. Common strategies consist of learning domain-invariant features~\cite{ganin,ltir}, often using adversarial learning in the process~\cite{ganin,fada,chen_no_2017,hoffman_fcns_2016,adaptsegnet}.
A popular trend in UDA is \textit{Self-Training}. These methods rely on self-supervision to learn from unlabelled data. In UDA, a successful strategy consists of leveraging target-curated pseudo-labels. Popular approaches for this purpose make use of confidence~\cite{cbst,iast,zou2019confidence}, try to balance the class predictions~\cite{zou2018unsupervised,hoyer2021daformer}, or use prototypes~\cite{chen2019progressive,zhang_category_2019,zhang_prototypical_2021} to improve the quality of the pseudo-labels.
Among many domain shifts, the synthetic-to-real one is the most studied, since the earliest works~\cite{cyclegan,dcan,cycada} to the latest~\cite{wu2022d2ada,lee2022bi,hoyer2022hrda,lai2022decouplenet,gong2022tacs,pan2022ml,jiang2022prototypical}. However, this shift is one of a kind since it occurs only once after training, and without the requirement of avoiding forgetting the source domain.
\begin{figure*}[t]
    \centering
    \includegraphics[width=0.95\textwidth]{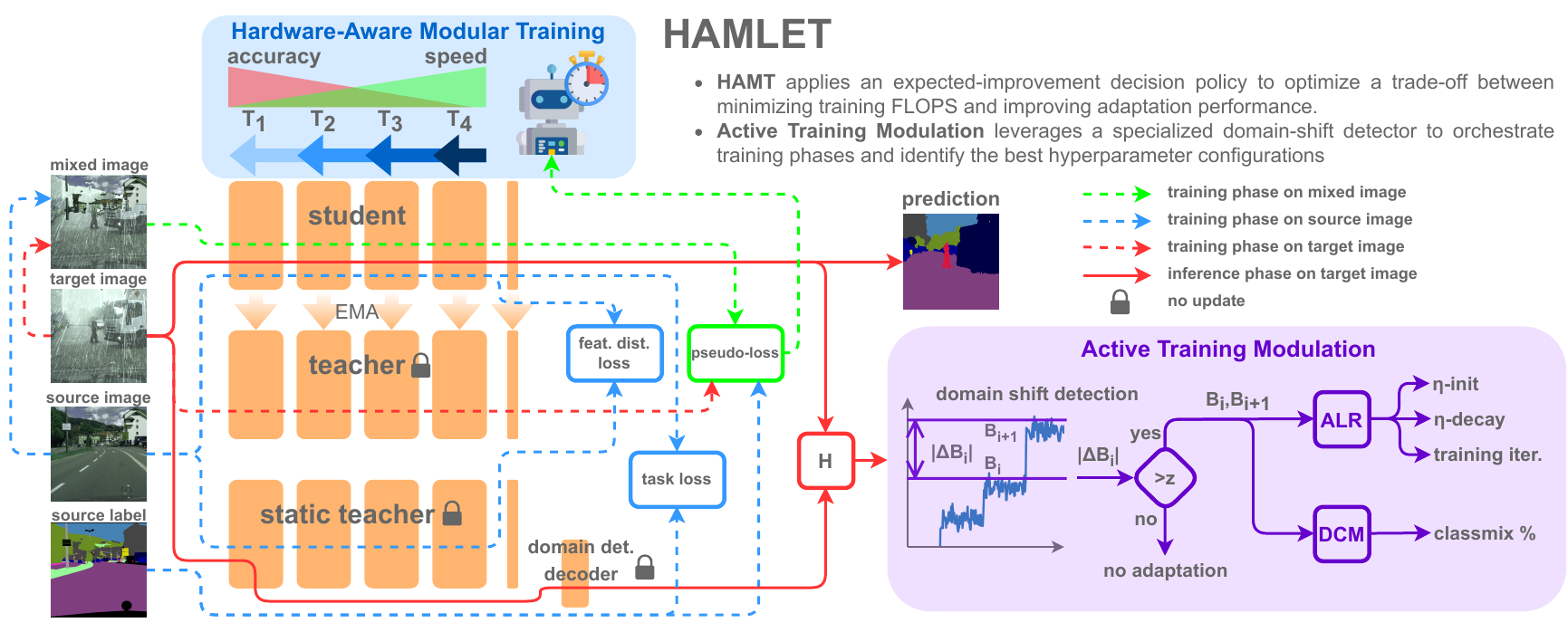}
    \vspace{-0.3cm}
    \caption{\textbf{HAMLET framework.} We employ a student-teacher model with an EMA and a static teacher. HAMT orchestrates the back-propagation over the student restricting it to a network subsection. The Active Training Modulation instead controls the adaptation process by selectively enabling it only when necessary as well as tweaking sensitive training parameters. }
    \label{fig:architecture}
\end{figure*}

\textbf{Continuous/Test-Time UDA.} This family of approaches marries UDA with continuous learning, thus dealing with the \textit{catastrophic forgetting} issue ignored in the synthetic-to-real case. 
Most \textit{continuous UDA} approaches deal with it by introducing a Replay Buffer~\cite{bobu_adapting_2018,lao2020continuous,kuznietsov2022towards}, while additional strategies make use of style transfer~\cite{wu_ace_2019}, contrastive~\cite{su_gradient_2020, vs2023towards} or adversarial learning~\cite{wulfmeier_incremental_2018}. Despite the definition, continuous UDA often deals with \textit{offline} adaptation, with well-defined target domains over which to adapt. Conceptually similar to it, is the branch of \textit{test-time adaptation}, or \textit{source-free} UDA, although tackling the problem in deployment rather than offline -- \ie{} with no access to the data from the source domain~\cite{Stan2021UnsupervisedMA}. Popular strategies to deal with it consist of generating pseudo-source data to avoid forgetting~\cite{liu_source-free_2021}, freezing the final layers in the model~\cite{liang_we_2020}, aligning features~\cite{liu2021ttt}, batch norm retraining through entropy minimization~\cite{wang2021tent} or prototypes adaptation~\cite{iwasawa2021testtime}.

\textbf{Online UDA.} Although similar in principle to test-time adaptation, online UDA~\cite{shift,Panagiotakopoulos_ECCV_2022,Volpi_2022_CVPR} aims to tackle multiple domain shifts, occurring unpredictably during deployment in real applications and without clear boundaries between them. On this track, the SHIFT dataset~\cite{shift} provides a synthetic benchmark specifically thought for this scenario, while OASIS~\cite{Volpi_2022_CVPR} proposes a novel protocol to evaluate UDA approaches, considering an online setting and constraining the evaluated methods to deal with frame-by-frame sequences.
As for methods, OnDA~\cite{Panagiotakopoulos_ECCV_2022} implements self-training as the orchestration of a static and a dynamic teacher to achieve effective online adaptation while avoiding forgetting, yet introducing massive overhead.

Real-time performance is an essential aspect of online adaptation, particularly in applications such as autonomous driving where slow models are impractical. A slow adaptation process not only limits the practicality of real-world applications but also fails to provide high accuracy until the adaptation is complete, thereby defeating the original purpose. Therefore, accelerating the adaptation process is crucial for achieving high accuracy in real-time scenarios. 

\section{Methods}
This section introduces HAMLET, a framework for \textbf{H}ardware-\textbf{A}ware \textbf{M}odular \textbf{L}east \textbf{E}xpensive \textbf{T}raining.
The framework aims to solve the problem of online domain adaptation with real-time performance through several synergistic strategies. First, we introduce a Hardware-Aware Modular Training (HAMT) agent able to optimize online a trade-off between model accuracy and adaptation time. HAMT allows us to significantly reduce online training time and GFLOPS. Nevertheless, the cheapest training consists of no training at all. Therefore, as the second strategy, we introduce a formal geometric model for online domain shifts that enable reliable domain shift detection and domain estimator signals (Adaptive Domain Detection, Sec.~\ref{subs:adaptive-domain-detection}). These can be easily integrated to activate the adaptation process only at specific times, \textit{as least as possible}. Moreover, we can further leverage these signals by designing adaptive training policies that dynamically adapt domain-sensitive hyperparameters. We refer to these as Active Training Modulations. We present an overview of HAMLET in Fig.~\ref{fig:architecture}.

\subsection{Model Setup}
Our approach builds on the recent progress in unsupervised domain adaptation and segmentation networks. We start with DAFormer~\cite{hoyer2021daformer}, a state-of-the-art UDA method, and adopt SegFormer~\cite{xie2021segformer} as our segmentation backbone due to its strong generalization capacity. We use three instances of the backbone, all pre-trained on the source domain: a student, a teacher, and a static (\ie{} frozen) teacher. During training, the student receives a mix of target and source images~\cite{dacs} and is supervised with a ``mixed-sample'' cross-entropy loss, $\mathcal{L}_T$ (represented by green, blue and red dashed lines, in Fig. \ref{fig:architecture}). This loss is computed by mixing the teacher's pseudo-labels and source annotations. To improve training stability, the teacher is updated as the exponential moving average (EMA) of the student.
To further regularize the student, we use source samples stored in a replay buffer and apply two additional losses (blue lines in Fig. \ref{fig:architecture}). First, we minimize the feature distance (Euclidean) between the student and the static teacher's encoder, $\mathcal{L}_{FD}$. Then, we employ a supervised cross-entropy task loss $\mathcal{L}_S$. Our complete objective is $\mathcal{L}=\mathcal{L}_S+\mathcal{L}_T+\lambda_{FD} \mathcal{L}_{FD}$, with $\lambda_{FD}$ being a weight factor. During inference on the target domain, only the student is used (red lines in Fig. \ref{fig:architecture}).

\subsection{Hardware-Aware Modular Training (HAMT)}
\label{subs:hardware-aware-modular-training}
Online adaptation requires updating the parameters during deployment time.
However, back-propagation is computationally expensive and hence too slow to be continuously applied on a deployed agent.
Opting for a partial weight update, for example by finetuning the last module of the network, would enable much more efficient training time.
However, domain shifts can manifest as changes in both the data input distribution (such as attributes of the images, \eg{} day/night) and the output distribution (\eg{} class priors).
This information could be encoded in different parts of the network, therefore just updating the very last segment might not suffice.
This motivates the need for orchestrating the training process, to ensure sufficient training while minimizing the computational overhead.
Inspired by reward-punishment~\cite{tonioni2019real} and reinforcement learning~\cite{wang2018haq} policies, we introduce an orchestration agent in charge of deciding how deeply the network shall be fine-tuned through a trade-off between the pseudo-loss minimization rate and the computational time.
In contrast to previous efficient back-propagation approaches~\cite{wei2017minimal, jiang2019accelerating, cheng2022stochastic}, our model is pre-trained on the task and thus requires smaller updates to adapt.
Let us start by modeling the problem. Our model backbone, $f$, is composed of four different modules: $f = m_4 \circ m_3 \circ m_2 \circ m_1$.
This defines our action space $\mathcal{A}=\{\mathrm{T_1}, \mathrm{T_2}, \mathrm{T_3}, \mathrm{T_4}\}$ where $\mathrm{T_4}$ corresponds to training just the last module of the network, $m_4$, while $\mathrm{T_3}$ the last two modules, \ie{} $m_4 \circ m_3$, $\mathrm{T_2}$ the last three, \ie{} $m_4 \circ m_3 \circ m_2$, and $\mathrm{T_1}$ the whole network $f$.
We also define a continuous state space $\mathcal{S}=\{\mathrm{R}, \mathrm{V}\}$ where $\mathrm{R}$ is the second derivative of the EMA teacher pseudo-loss, $l_t$, over time, hence $\mathrm{R_t}=-\frac{\Delta^2 l}{(\Delta t)^2}$, computed in discrete form as $R_t=-(l_t - 2l_{t-1} + l_{t-2})$. $\mathrm{V}$ represents a cumulative vector with the same dimension as the action space $\mathcal{A}$, initialized at zero. Now we have everything in place to employ an expected-improvement based decision model. 
At each time-step $t$, action $T_j$ is selected for $j = \operatorname{argmax} \mathrm{V}_t$. During training step $\mathrm{t}$, $\mathrm{V}[j]$ is updated as:
{
\begin{equation}
\label{eq:naive_update}
\mathrm{V}[j]_{t+1} = \alpha R_t  + (1 - \alpha) \mathrm{V}[j]_{t}  
\end{equation}
}
where $\alpha$ is a smoothing factor, \textit{\eg{}} $0.1$. \ie$\mathrm{V_{t}}$ hold a discrete exponential moving average of $R_t$. Therefore, our policy can be seen as a greedy module selection based on the highest expected loss improvement over its linear approximation. A notable drawback of this policy is that we will inevitably converge towards picking more rewarding, yet expensive, actions \ie{} $\mathrm{T_1}, \mathrm{T_2}$ compared to more efficient but potentially less effective actions \ie{} $\mathrm{T_3}, \mathrm{T_4}$. However, our goal is not to maximize $ -\frac{\Delta^2 l}{(\Delta t)^2}$ where $\Delta t$ is the number of updates, our goal is instead to maximize $ -\frac{\Delta^2 l}{(\Delta \tau)^2}$ where $\Delta \tau$ is a real-time interval. Therefore, we have to introduce in the optimization policy some notion of the actual training cost of each action in $\mathcal{A}$ on the target device.
To start with, we measure the training time associated with each action, obtaining $\omega_T = \{ \omega_{T_1}, \omega_{T_2}, \omega_{T_3}, \omega_{T_4} \}$. With this we can compute the time-conditioning vector $\gamma$ as
{
\small
\begin{equation}
\gamma_j=\frac{e^{\frac{1}{\beta\omega_{T_j}} }}{\sum_{k=1}^K e^{\frac{1}{\beta\omega_{T_k}}}}
\quad
\text{ for } j=1, \ldots, K
\end{equation}
}
where $\beta$ is the softmax temperature, and $K$ the number of actions, \ie{} 4 in our model.
We modify our update policy to favor less computationally expensive modules by scaling the updates with $\gamma$, replacing Eq.~\ref{eq:naive_update} with:

{
\small
\begin{equation}
\mathrm{V}[j]_{t+1} = 
 \begin{cases}
 \gamma_j \alpha R_t + (1 - \alpha) \mathrm{V}[j]_{t}  & \textrm{if } R_t \geq 0\\
 (1 - \gamma_j) \alpha R_t  + (1 - \alpha) \mathrm{V}[j]_{t} & \textrm{if } R_t < 0\\
 \end{cases}
\end{equation}
}

This policy makes it so that more expensive actions receive smaller rewards and larger punishments.
Despite its simplicity, this leads to a significant reduction in FLOPS for an average back-propagation $\beta$, \ie{} $-30\%$ with $\beta=2.75$ or $-43\%$ with $\beta=1$. We finally choose $\beta=1.75$ to obtain a FLOPS reduction of $-34\%$. Exhaustive ablations on HAMT are presented in the {supplementary material}. 

\subsection{Active Training Modulation}
\label{subs:active-training-modulation}
Continuous and test-time adaptation methods tackle online learning as a continuous and constant process carried out on the data stream. Nevertheless, this approach presents several shortcomings when it comes to real-world deployments. Performing adaptation when the deployment domain is unchanged does not lead to further performance improvements on the current domain; instead, it might cause significant forgetting on previous domains, hence hindering model generalization (we present evidence of this in the {supplementary material}). Even if mitigated by HAMT, online training remains a computationally expensive procedure, also due to several teachers' necessary forward passes.
However, knowing when and what kind of adaptation is needed is not a trivial task. We tackle this by introducing an Adaptive Domain Detection mechanism, in Sec.~\ref{subs:adaptive-domain-detection}, and then a set of strategies to reduce the training time while optimizing the learning rate accordingly, in Sec.~\ref{subs:learning-rate-scheduling}.

\subsubsection{Adaptive Domain Detection}
\label{subs:adaptive-domain-detection}
A key element of an online adaptation system consists of acquiring awareness of the trajectory in the data distribution space, \ie{} domains, traveled by the student model during deployment. We can model the problem by setting the trajectory origin in the source domain. With high dimensional data, the data distribution is not tractable, therefore the trajectory cannot be described in closed form. Recent work~\cite{Panagiotakopoulos_ECCV_2022} introduced the notion of distance between the current deployed domain and source by approximating it with the confidence drop of a source pre-trained model. This approach heavily relies on the assumption that the pre-trained model is well-calibrated. While this might hold for domains close to source, the calibration quickly degrades in farther domains~\cite{shift, Panagiotakopoulos_ECCV_2022}. This myopic behavior dampen the simple use of confidence for domain detection. Furthermore, the additional forward pass increases the computational cost during deployment.
We tackle these limitations with an equivalently simple, yet more robust, approach. We modify the backbone of the static teacher $f^\text{st}$ used for the feature distance loss $\mathcal{L}_{FD}$ by connecting a lightweight segmentation head, $d^\text{st}_{1}$, after the first encoder module $m_{1}^\text{st}$: $h^\text{st}_{1} = d^\text{st}_{1} \circ m^\text{st}_{1}$.
This additional decoder, $h^\text{st}_{1}$, is trained offline, on source data, without propagating gradients in the backbone ($m^\text{st}_{1}$ is frozen).
Given a target sample $x_T$, we propose to compute the cross-entropy between the one-hot encoded student prediction $p(x_T) = 1_{\operatorname*{argmax}(f(x_T))}$ and the lightweight decoder prediction $g(x_T) = h^\text{st}_{1}(x_T)$ as

\begin{equation}
H_T^{(i)}=-\sum_{p=1}^{H \times W} \sum_{c=1}^C p\left(x_T^{(i)}\right) \left.\log g\left(x_T^{(i)}\right)\right|_{p,c}
\end{equation}

Thanks to the student model's higher generalization capability (both due to a larger number of parameters and the unsupervised adaptation process), it will always outperform the lightweight decoder head.
Nevertheless, since now the distance is measured in the prediction space, we are not subjected to model miscalibration.
Furthermore, since the student model is in constant adaptation, the domain distance accuracy actually improves over time, leading to better results.
We present evidence of these claims in the {supplementary material}.
We now define a denoised signal by using bin-averaging $A_T^{(i)}=\sum_{j=mi}^{m(i+1)-1}\frac{H_T^{(j)}}{m}$ where $m$ is the bin size.
Domains are modeled as discrete steps of $A_T^{(i)}$

{
\small
\begin{equation}
    B_0 = A_0 \qquad
    B_i =
    \begin{cases}
        A_i & \textrm{if  $|B_{i-1} - A_i|>z$} \\
        B_{i-1} & \textrm{otherwise}
    \end{cases}
\end{equation}
}
where $B$ is the discretized signal and $z$ is the minimum distance used to identify new domains. Finally, we refer to the signed amplitude of domain shifts as $\Delta B_i = B_i - B_{i-1}$, and a domain change is detected whenever $|\Delta B_i| > z$.

\begin{table*}[t]
\scriptsize
\centering
\renewcommand{\tabcolsep}{12pt}
\scalebox{0.8}{
\begin{tabular}{c}
\begin{tabular}{ccccccr@{\tiny$\,\pm\,$}lr@{\tiny$\,\pm\,$}lr@{\tiny$\,\pm\,$}lr@{\tiny$\,\pm\,$}lr@{\tiny$\,\pm\,$}l@{~~}r@{\tiny$\,\pm\,$}l@{~~}r@{\tiny$\,\pm\,$}l@{~~}r@{\tiny$\,\pm\,$}l}
\hline
&      &    &     &     &     & \mc2c{ \cellcolor{orange!40} 200mm}           & \mc2c{\cellcolor{orange!40} All-domains}                 & \mc2c{\cellcolor{green!40}}          & \mc6c{\cellcolor{blue!25}Average GFLOPS} & \mc4c{\cellcolor{blue!25}Adaptation GFLOPS}   \\
& HAMT & LT & ALR & DCM & RCS & \mc2c{\cellcolor{orange!40} (mIoU)} & \mc2c{\cellcolor{orange!40} (mIoU)} & \mc2c{\cellcolor{green!40} FPS} & \mc2c{\cellcolor{blue!25}Total} & \mc2c{\cellcolor{blue!25}Fwd.}   & \mc2c{\cellcolor{blue!25}Bwd.}      & \mc2c{\cellcolor{blue!25}Fwd.} & \mc2c{\cellcolor{blue!25}Bwd.}  \\
\hline 
(A) & --         & --         & --         & --         & --         & 62.2 & {\tiny 0.9} & 69.5 & {\tiny 0.3} & 5.9 & {\tiny 0.0 } & 125.2 & {\tiny 0.0} & 94.4 & {\tiny 0.0} & 30.8 & {\tiny 0.0} & 56.6 & {\tiny 0.0} & 30.8 & {\tiny 0.0} \\
(B) & \checkmark & --         & --         & --         & --         & 60.2 & {\tiny 0.5} & 68.7 & {\tiny 0.3} & 7.0 & {\tiny 0.1 } & 114.7 & {\tiny 0.0} & 94.4 & {\tiny 0.0} & 20.3 & {\tiny 0.0} & 56.6 & {\tiny 0.0} & 20.3 & {\tiny 0.0} \\
(C) & \checkmark & \checkmark & --         & --         & --         & 51.8 & {\tiny 0.5} & 65.7 & {\tiny 0.2} & 29.5 &{\tiny  0.6} &  44.4 & {\tiny 0.5} & 42.6 & {\tiny 0.4} & 1.8  & {\tiny 0.2} & 56.6 & {\tiny 0.0} & 20.2 & {\tiny 0.2}  \\
(D) & \checkmark & \checkmark & \checkmark & --         & --         & 54.1 & {\tiny 1.2} & 65.9 & {\tiny 0.2} & 29.5 &{\tiny  0.5} &  44.4 & {\tiny 0.3} & 42.7 & {\tiny 0.2} &  1.8 & {\tiny 0.1} & 56.6 & {\tiny 0.0} & 20.3 & {\tiny 0.1}  \\
(E) & \checkmark & \checkmark & \checkmark & \checkmark & --         & 56.6 & {\tiny 0.8} & 66.3 & {\tiny 0.1} & 28.9 &{\tiny  0.3} &  44.7 & {\tiny 0.2} & 42.9 & {\tiny 0.2} & 1.8  & {\tiny 0.1} & 56.6 & {\tiny 0.0} & 20.2 & {\tiny 0.0}  \\
(F) & \checkmark & \checkmark & \checkmark & --         & \checkmark & 55.8 & {\tiny 1.0} & 66.3 & {\tiny 0.2} & 29.1 &{\tiny  1.1} &  45.2 & {\tiny 0.1} & 43.2 & {\tiny 0.1} & 2.0  & {\tiny 0.0} & 56.6 & {\tiny 0.0} & 20.3 & {\tiny 0.0}  \\
(G) & \checkmark & \checkmark & \checkmark & \checkmark & \checkmark & 58.2 & {\tiny 0.8} & 66.9 & {\tiny 0.3} & 29.7 &{\tiny  0.6} &  45.7 & {\tiny 0.3} & 43.6 & {\tiny 0.2} & 2.1  & {\tiny 0.1} & 56.6 & {\tiny 0.0} & 20.2 & {\tiny 0.1}  \\
\hline
\end{tabular}
\\ (a) 
\\
\begin{tabular}{lrrrrrrrrrrrr}
\toprule
{} &  clear 1 &  200mm &  clear 2 &  100mm &  clear 3 &  75mm &  clear 4 &  \cellcolor{orange!40} clear h-mean & \cellcolor{orange!40} target h-mean & \cellcolor{orange!40} total h-mean &  \cellcolor{green!40}  FPS & \cellcolor{blue!25} GFLOPS\\
\midrule
(A) &   72.9 &   52.2 &     73.6 &   64.2 &     73.0 &  67.6 &     73.4 &         73.2 &          60.6 &       67.2 &  5.6 &   125.2 \\
(B)       &   73.0 &   50.4 &     73.4 &   62.1 &     73.0 &  67.3 &     73.2 &         73.1 &          59.1 &       66.4 &  6.8 &   114.7 \\
(C)        &   73.4 &   46.0 &     73.5 &   61.5 &     73.6 &  66.1 &     73.8 &         73.6 &          56.5 &       65.1 &  7.2 &   100.0 \\
(G)       &   73.4 &   53.6 &     73.1 &   65.2 &     73.5 &  68.2 &     73.2 &         73.3 &          61.6 &       67.8 &  9.1 &    82.2 \\
\bottomrule
\end{tabular} \\ (b) \\
\end{tabular}}

\vspace{-0.3cm}
\caption{\textbf{Ablation studies -- HAMLET components.} Top: Increasing Storm (8925 frames per domain) \cite{Panagiotakopoulos_ECCV_2022}, bottom: Fast Storm C~\cite{Panagiotakopoulos_ECCV_2022} (2975 frames per domain). For each configuration, we report mIoU, framerate, and GFLOPS.}
\vspace{-0.35cm}
\label{tab:ablation}
\end{table*}

\subsubsection{Least Training and Adaptive Learning Rate}
\label{subs:learning-rate-scheduling}
The definitions of $B$ allow us to customize the training process. To this end, we adopt a \textit{Least Training} (LT) strategy and trigger adaptation only when facing a new domain, which occurs when $|\Delta B_i| > z$. Effective online learning performance depends heavily on the choice of hyperparameters such as the learning rate $\eta$ and learning rate decay rate. Therefore, we can adjust these parameters to facilitate adaptation according to the nature and intensity of domain shifts we encounter, we refer to this orchestration as Adaptive Learning Rate (ALR). For example, the larger the domain shift (\ie $|\Delta B_i|$), the more we need to adapt to counteract its effect. This can be achieved by either running more optimization steps or using a higher learning rate.
Whenever a domain shift is detected, we compute the number of adaptation iterations $L = K_l\frac{|\Delta B_i|}{z}$, hence proportionally to the amplitude of the shift $|\Delta B_i|$ relative to the threshold $z$. $K_l$ is a multiplicative factor representing the minimum adaptation iterations. If a new domain shift takes place before the adaptation process completes, we accumulate the required optimization steps.
Then, we can play on two further parameters: $K_l$ and the learning rate schedule. We argue that proper scheduling is crucial for attaining a smoother adaptation. The learning rate, $\eta$, is linearly decayed until the adaptation is concluded -- the smaller the domain shift, the faster the decay. While the initial learning rate, $K_\eta$, should be higher when the domain shift is triggered in domains farther from the source

{
\small
\begin{align}
    K_\eta & = K_{\eta,\textrm{min}}+\frac{(B_\textrm{i}-B_\textrm{source})(K_{\eta,\textrm{max}}-K_{\eta,\textrm{min}})}{B_\textrm{hard}-B_\textrm{source}}
\end{align}
}
where $B_{\text{source}}$ (resp. $B_{\text{hard}}$) is an estimate of $B$ when the network is close to (resp. far from) the source domain; and $K_{\eta,\text{min}}$ (resp. $K_{\eta,\text{max}}$) is the value of $K_\eta$ assigned when the network is close to (resp. far away from) the source. Concerning $K_l$, we posit that moving towards the source requires less adaptation than going towards harder domains: the model shows good recalling of previously explored domains and thanks to the employed regularization strategies

{
\small
\begin{equation}
    K_l =
    \begin{cases}
        K_{l,\text{max}} & \textrm{if $\Delta B_i \geq 0$} \\
        K_\textrm{l,min}+\frac{(B_\textrm{i}-B_\textrm{source})(K_\textrm{l,max}-K_\textrm{l,min})}{B_\textrm{hard}-B_\textrm{source}} & \textrm{otherwise}
    \end{cases}
\end{equation}
}
where $K_{l,\text{min}}$ (resp. $K_{l,\text{max}}$) is the value of $K_l$ assigned when the model is close to (resp. far away from) the source domain. Extensive ablations in the {supplementary material} will highlight how the orchestration of the adaptation hyper-parameters improves the accuracy-speed trade-off.

\subsubsection{Dynamic ClassMix (DCM)}

ClassMix~\cite{olsson2021classmix} provides a simple mechanism for data augmentation by mixing classes from the source dataset into target images.
Usually 50\% of the classes in the source dataset are selected, however we notice that this percentage is a highly sensitive hyperparameter in online domain adaptation.
Injecting a significant portion of source classes has a beneficial impact when adapting to domains closer to the source domain, whereas when adapting to domains further from the source the opposite effect can be observed, as it effectively slows down the adaptation process.
We therefore exploit once more the deployment domain awareness to control the mixing augmentation:

{
\footnotesize
\begin{equation}
    K_\textrm{CM} = K_{\text{CM},\text{min}} + \frac{(B_\textrm{i}-B_\textrm{source})(K_\textrm{CM,max}-K_\textrm{CM,min})}{B_\textrm{hard}-B_\textrm{source}}.
\end{equation}
}
where $K_\textrm{CM}$ is the percentage of source classes used during adaptation; and $K_\textrm{CM, min}$ (resp. $K_\textrm{CM, max}$) is the value of $K_\textrm{CM}$ assigned when the network is close to (resp. far away from) the source domain.

\subsubsection{Buffer Sampling}
Following~\cite{Panagiotakopoulos_ECCV_2022}, to simulate real deployment, we limit our access to the source domain by using a replay buffer. Additionally, instead of initializing at random (with a uniform prior), we apply Rare Class Sampling (RCS) (skewed priors) as in~\cite{hoyer2021daformer}. This incentives a more balanced class distribution over the buffer, ultimately leading to better accuracy.

\section{Experimental Results}
\begin{figure}[t]
    \centering
    \includegraphics[width=0.4\textwidth]{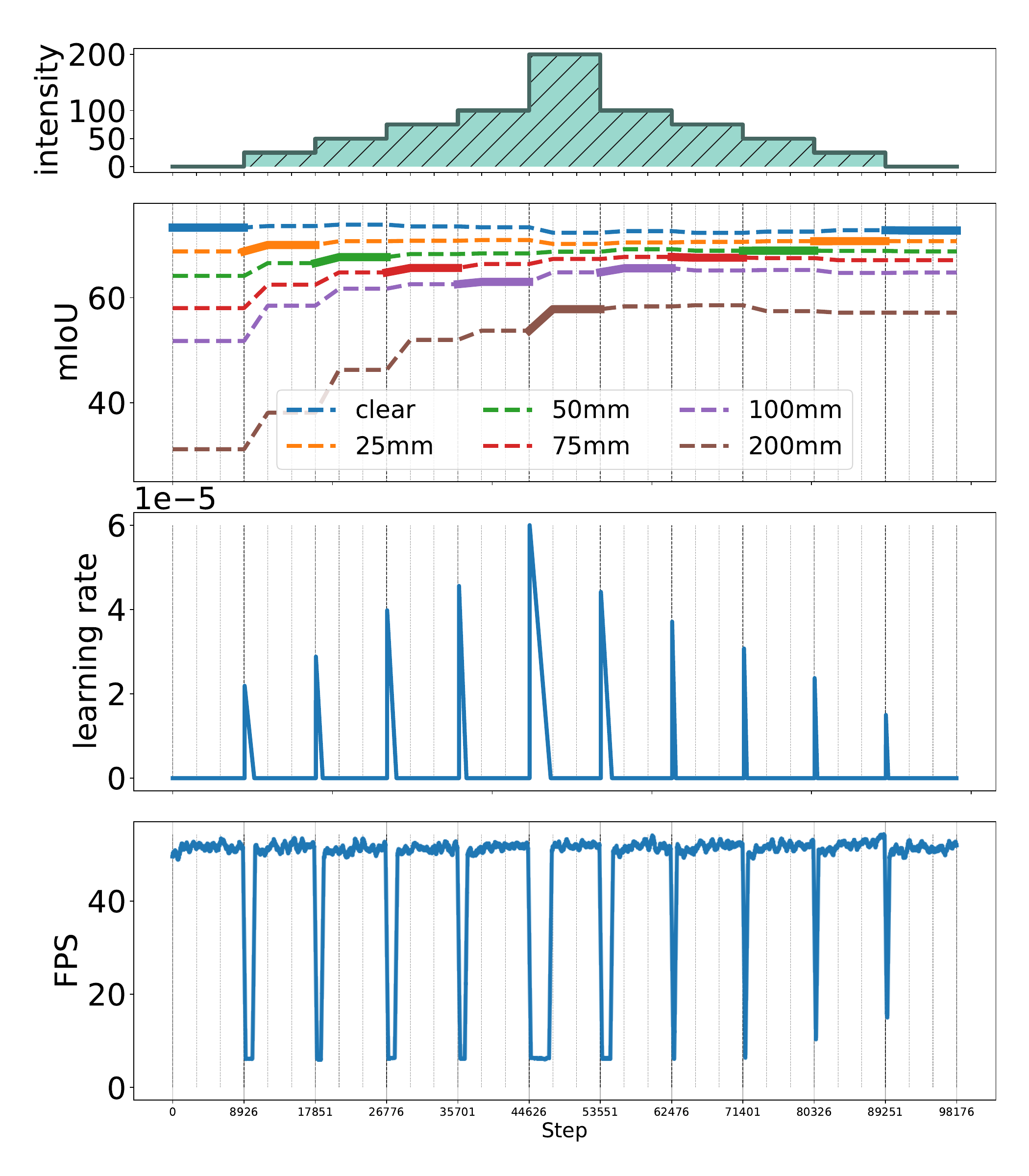}
    \vspace{-0.5cm}
    \caption{\textbf{HAMLET on the Increasing Storm.} We show rain intensity (in millimetres), mIoU over active (bold) and inactive (dashed) domains, learning rate and FPS.}
    \vspace{-0.3cm}
    \label{fig:increasing_storm}
\end{figure}

The experiments are carried out on (a) the OnDA benchmarks~\cite{Panagiotakopoulos_ECCV_2022} and (b) the SHIFT dataset~\cite{shift}. (a) is a semi-syntehtic benchmark, as it applies synthetic rain and fog~\cite{tremblay2020rain} over 4 different intensities profiles. The main benchmark, Increasing Storm, presents a storm with a pyramidal intensity profile; see Fig.~\ref{fig:increasing_storm}. In contrast, (b) is a purely synthetic dataset, where both the underlying image and the weather are synthetically generated and thus domain change is fully controllable. 
All models are evaluated using mIoU: following~\cite{Panagiotakopoulos_ECCV_2022}, we report the harmonic mean over domains to present the overall adaptation performance. All experiments were carried out using an Nvidia\texttrademark~RTX~3090 GPU. We refer to {supplementary material} for further details.
\newcommand*{\cdash}{\multicolumn1c{--}}
\begin{table*}[t]
    \centering
    \scalebox{0.65}{
    \begin{tabular}{clrr rr rr rr rr r rrrrr}
    \toprule
    {} & & \mc{2}{c}{clear} & \mc{2}{c}{25mm} & \mc{2}{c}{50mm} & \mc{2}{c}{75mm} & \mc{2}{c}{100mm} &  200mm &  \mc{3}{c}{\cellcolor{orange!40}  h-mean} &      \cellcolor{green!40}  {FPS}   & \cellcolor{blue!25}  {GFLOPS}\\
    & {} &         F &          B &      F &          B &          F &          B &      F &          B &          F &          B &      F &         \cellcolor{orange!40}  F & \cellcolor{orange!40} B & \cellcolor{orange!40}  T & \cellcolor{green!40}  & \cellcolor{blue!25} \\
    \midrule
    (A) & DeepLabV2 (no adaptation) & 64.5 & \cdash & 57.1 & \cdash & 48.7 & \cdash & 41.5 & \cdash & 34.4 & \cdash & 18.5 & 37.3 & \cdash & \cdash & 39.4 &  \cdash\\
    \midrule
    (B) & DeepLabV2 fully supervised (oracle) &  {64.5} &\cdash  & {64.1} &\cdash  & {63.7} &\cdash &  {63.0} &\cdash &  {62.4} &\cdash &   {58.2} & {62.6} &\cdash&\cdash& 39.4 &\cdash\\
    \midrule
    (C) & OnDA & 64.5 & 64.8 & 60.4 & 57.1 & 57.3 & 54.5 & 54.8 & 52.2 & 52.0 & 49.1 & 42.2 & 54.2 & 55.1 &\cdash& 1.3 & \cdash\\
    
    \midrule
    \midrule
    (D) & SegFormer MiT-B1 (no adaptation) &  73.4 & \cdash&  68.8 & \cdash&  64.2 & \cdash&  58.0 & \cdash&  51.8 & \cdash&  31.2 &      57.8 &  \cdash& \cdash&   48.4 & 34.9 \\ 
    (E) & SegFormer MiT-B5 (no adaptation) &  77.6 & \cdash&  73.9 & \cdash&  71.0 & \cdash&  67.2 & \cdash&  62.6 & \cdash&  46.7 & 64.7 &  \cdash& \cdash&   11.5 & 240.4  \\ 
        \midrule
    (F) & SegFormer MiT-B1 fully supervised (oracle)  &  72.9 & \cdash&  72.4 & \cdash&  72.1 & \cdash&  71.5 & \cdash&  70.7 & \cdash&  68.6 & 71.3 &  \cdash& \cdash&   48.4 & 34.9 \\
    \hline 
    (G) & TENT            &  73.0 &  72.8 &  68.5 &  68.6 &  64.5 &  64.8 &  59.7 &  60.2 &  54.5 &  54.8 &  35.9 &   56.2 &  63.6 &  59.9 &  10.0 &\cdash\\
    (H) & TENT + Replay Buffer &  73.0 &  72.8 &  68.5 &  68.6 &  64.5 &  64.8 &  59.7 &  60.2 &  54.4 &  54.7 &  35.8 &   56.1 &  63.6 &      59.9 &   7.8 &\cdash\\
    (I) & CoTTA           &  72.5 & 74.4 & 69.5 & \bfseries 70.9 & 65.9 & \bfseries 68.2 & 66.1 & 64.7 & 64.6 & 63.5 & 57.2 & 65.6 & \bfseries 68.1 & 66.8 & 0.6 & 593.8 \\
    (J) & CoTTA \textit{real-time} & 73.3 & \bfseries 75.4 & \bfseries 70.3 & 70.6 & 66.9 & 66.4 & 62.5 & 61.4 & 57.6 & 56.9 & 39.7 & 59.2 & 65.5 & 62.3 & 27.0 & \bfseries 41.7 \\
    (K) & HAMLET (ours)            &  \bfseries 73.4 &  71.0 & 70.1 &  68.8 &  \bfseries 67.7 &  67.5 &  \bfseries 66.6 &  \bfseries 66.4 &  \bfseries 65.5 &  \bfseries 64.6 &  \bfseries 59.2 &  \bfseries 66.8 &  67.6 &  \bfseries 67.2 &  \bfseries 29.1 & 45.7  \\
    \bottomrule
    \end{tabular}
    }
    \vspace{-0.3cm}
    \caption{\textbf{Comparison against other models -- Increasing storm scenario.} (A-C) methods built over DeepLabv2, (D-E) SegFormer variants trained on source, (F) oracle, (G-K) models adapted online. We report mIoU, framerate, and GFLOPS.}
    \label{tab:comparison}
\end{table*}

\subsection{Ablation Studies}

In Tab.~\ref{tab:ablation} we study the impact of each contribution to adaptation performance, both in terms of accuracy and efficiency. 
For each configuration, we report mIoU over different portions of the sequence, the framerate and the amount of GFLOPS -- respectively averages of: total, forward and backward passes, and dedicated adaptation only, also divided in forward (Fwd) and backward (Bwd). 
Tab.~\ref{tab:ablation} (a) shows results on the Increasing Storm scenario~\cite{Panagiotakopoulos_ECCV_2022}. Here, we show mIoU over the 200mm domain, \ie{} the hardest in the sequence, as well as the mIoU averaged over forward and backward adaptation, \ie, from \textit{clear} to 200mm rain and backward.
Results are averaged over 3 runs with different seeds, with standard deviation being reported. (A) reports the results achieved by na\"ively performing full adaptation of the model. HAMT can increase the framerate by roughly 15\% by reducing the Bwd GFLOPS of 34\%, at the expense of as few as 0.7 mIoU on average, \ie, about 2 points on the 200mm domain. 
The main boost in terms of speed is obviously given by LT (C), which inhibits the training in absence of detected domain shifts. LT increases the framerate by approximately $4\times$ by decimating the total GFLOPS, yet not affecting the adaptation Bwd GFLOPS. This comes with a price in terms of mIoU, dropping by about 4 points on average and more than 10 points on 200mm -- not a moderate drop anymore. LT impact highly depends on the domain sequence experienced during deployment: frequent domain changes could prevent training inhibition, thus neglecting LT gains in terms of efficiency, as we will appreciate later.
The loss in accuracy is progressively regained by adding ALR (D), with further improvements yielded by one between DCM (E) and RCS (F), or both together (G) leading to the full HAMLET configuration. The three together allow for reducing the gap to 2.5 points mIoU -- 4 over the 200mm domain -- without sacrificing any efficiency. 
Tab.~\ref{tab:ablation} (b) shows further results, on a faster version of Storm C~\cite{Panagiotakopoulos_ECCV_2022}. This represents a much more challenging scenario, with harsher and $3\times$ more frequent domain shifts. Here we show the single domains mIoU, as well as harmonic mean on source and target domains, and all frames. As expected, in this benchmark, LT alone (C) results much less effective than before, with a much lower gain in FPS and GFLOPS. Here, the synergy between the HAMT, LT, and the other components (G) allows for the best accuracy and speedup -- even outperforming the full training variant (A) -- highlighting their complementarity.
Further ablations are in the {supplementary material}.

\begin{table*}[t]
\scriptsize
    \centering
    \scalebox{0.78}{
    \renewcommand{\tabcolsep}{10pt}
    \begin{tabular}{lrrrrrrrrrrrrrrr}
    \toprule
    {} & \mc{2}{c}{clear} & \mc{2}{c}{750m} & \mc{2}{c}{375m} & \mc{2}{c}{150m} &  75m & \mc{3}{c}{\cellcolor{orange!40}h-mean} &      \cellcolor{green!40} FPS &  \cellcolor{blue!25}GFLOPS  \\
    {} &          F &          B &      F &          B &          F &          B &      F &          B &          F &          \cellcolor{orange!40}F & \cellcolor{orange!40}B & \cellcolor{orange!40}T & \cellcolor{green!40} & \cellcolor{blue!25}\\
    \midrule  
    OnDA & 64.9 & 65.8 & 63.3 & 62.3 & 60.7 & 58.8 & 51.6 & 49.1 & 42.1 & 55.1 & 54.1 & -- & 1.3 & -- \\
    \hline
    SegFormer MiT-B1 (no adaptation)   &  71.1 &  -- &  70.0 &  -- &  67.5 &  -- &  58.8 &  -- &  46.9 &  61.3 &  -- & -- &   48.4 & 34.9 \\
    Full training &  71.5 &  72.1 &  72.9 &  74.7 &  71.9 &  73.1 &  67.6 &  68.1 &  61.3 &   68.7 &  71.9 &      70.3 &   5.6 & 125.2 \\
    HAMLET (ours) &  71.1 &  71.6 &  70.3 &  70.8 &  68.8 &  69.2 &  64.3 &  64.3 &  57.0 &   65.9 &  68.9 &      67.4 &  24.8 & 50.7 \\
    \bottomrule
    \end{tabular}
    }
    \vspace{-0.3cm}
    \caption{\textbf{Results on foggy domains.} Comparison between OnDA, Source SegFormer, full training adaptation, and HAMLET.}
    \vspace{-0.3cm}
    \label{tab:fog}
\end{table*}
\begin{table*}[t]
\scriptsize
    \centering
    \scalebox{0.78}{
    \renewcommand{\tabcolsep}{8pt}
    \begin{tabular}{lrrrrrrrrrrrrrrrr}
    \toprule
    {} & \mc{2}{c}{Clear} & \mc{2}{c}{Cloudy} & \mc{2}{c}{Overcast} & \mc{2}{c}{Small rain} & \mc{2}{c}{Mid rain} &  Heavy rain & \mc{3}{c}{\cellcolor{orange!40}h-mean} &      \cellcolor{green!40} FPS   & \cellcolor{blue!25}GFLOPS\\
    {} &          F &          B &      F &          B &          F &          B &      F &          B &          F &          B &      F &          \cellcolor{orange!40}F & \cellcolor{orange!40}B & \cellcolor{orange!40}T & \cellcolor{green!40} &\cellcolor{blue!25}\\
    \midrule
    SegFormer MiT-B1 fully supervised (oracle)  &  80.1 &  -- &   79.9 &  -- &     79.8 &  -- &       78.9 &  -- &     78.7 &  -- &       77.1 &   79.1 &  -- &    -- &   48.4 &    34.93 \\
    \hline 
    SegFormer MiT-B1 (no adaptation)  &  79.6 &  -- &   77.1 &  -- &     75.4 &  -- &       73.4 &  -- &     71.4 &  -- &       66.7 &   73.7 &  -- &    -- &   48.4 &    34.93 \\
    Full training &  78.9 &  79.3 &   76.7 &  76.8 &     76.8 &  77.9 &       74.8 &  74.8 &     76.3 &  76.5 &       74.0 &   76.2 &  77.0 &      76.6 &   5.0 &  125.1 \\
    HAMLET (ours)  &  79.6 &  78.9 &   76.9 &  76.6 &     76.1 &  77.4 &       73.3 &  74.3 &     74.2 &  76.0 &       74.2 &   75.7 &  76.6 &      76.1 &  26.8 &   43.9 \\
    \bottomrule
    \end{tabular}}
    \vspace{-0.3cm}
    \caption{\textbf{Results on SHIFT dataset~\cite{shift}.} Comparison between Source SegFormer, full training adaptation, and HAMLET.}
    \label{tab:shift}
\end{table*}

\subsection{Results on Increasing Storm}
Tab.~\ref{tab:comparison} shows a direct comparison between HAMLET and relevant approaches. The presented test-time adaptation strategies namely -- TENT and CoTTA -- were revised to handle the online setting and be fairly compared with HAMLET. All methods start with the same exact initial weights -- with HAMLET requiring the additional lightweight decoder, not needed by TENT and CoTTA -- using SegFormer MiT-B1 as the backbone, since it is $4\times$ faster than SegFormer MiT-B5 and thus better suited to keep real-time performance even during adaptation.
We report results achieved by DeepLabv2 trained on source data only (A), an \textit{oracle} model trained with full supervision (B), as well as OnDA~\cite{Panagiotakopoulos_ECCV_2022} (C) as a reference.
Then, we report SegFormer models trained on the source domain only (D) and (E). In (F) we show the performance achieved by an oracle SegFormer, trained on all domains fully supervised.
Following~\cite{Panagiotakopoulos_ECCV_2022}, columns ``F'' concern forward adaptation from \textit{clear} to 200mm, while columns ``B'' show backward adaptation from 200mm to \textit{clear}, while the h-mean T refers to the overall harmonic mean. 
We can notice how SegFomer results are much more robust to domain changes with respect to DeepLabv2. Indeed, SegFormer MiT-B5 (E), without any adaptation, results more accurate than DeepLabv2 oracle (B), as well as better and faster than OnDA (C). The faster variant (D) outperforms OnDA both in speed and accuracy, reaching 48 FPS.
Nevertheless, domain changes still dampen the full potential of SegFormer. Indeed, the oracle (F) outperforms (D) by about +14 mIoU. However, this is not meaningful for real deployment experiencing unpredictable domain shifts, as it assumes to have data available in advance.
Concerning test-time models, TENT starts adapting properly only beyond 50mm, both with (G) and without (H) frame buffer, while it loses some accuracy on 25mm. This makes its overall forward adaptation performance slightly worse compared to the pre-trained model (D), while being better at backward adaptation. Despite outperforming SegFormer MiT-B1, TENT is both slower and less accurate than SegFormer MiT-B5 running without any adaptation, further suggesting the robustness of the latter and making TENT not suitable for real-world deployment.
On the contrary, CoTTA (I) outperforms both SegFormer models trained on source only, at the expense of dropping the framerate below 1FPS.
It is worth mentioning that these metrics were collected after each domain was completed by each model individually. In an evaluation setup imposing a shared time frame, slower models would present much lower metrics, since their adaptation process would result constantly lagged. In fact, forcing CoTTA to run in real-time, at nearly 30FPS -- \ie{} by training once every 50 frames -- dramatically reduces the effectiveness of the adaptation process (J), with drastic drops in the hardest domains.
Finally, HAMLET (K) succeeds on any fronts, improving the baseline (D) by about 10 points with only a cost of 25\% in terms of speed, while outperforming SegFormer MiT-B5 (E) both on accuracy (+2.5 mIoU) and speed ($3\times$ faster) -- being the only method achieving this, and thus the only suitable choice for real-time applications.
Fig.~\ref{fig:increasing_storm} shows the overall behavior of HAMLET while adapting over the Increasing Storm. In addition to the rain intensity and the mIoU achieved on each domain -- active (bold) or inactive (dashed), \ie{} respectively the mIoU on the domain being currently faced during deployment, and how the current adaptation affects the performance on the other domains to highlight the robustness to forgetting -- we also report how the learning rate is modulated in correspondence of detected domain shifts, with a consequent drop in FPS due to the short training process taking place. For further experiments on harsher and sudden adaptation cycles, we include results of Storms A, B, C~\cite{Panagiotakopoulos_ECCV_2022} in the {supplementary material}.

\begin{figure}[t]
    \centering
    \includegraphics[width=0.4\textwidth]{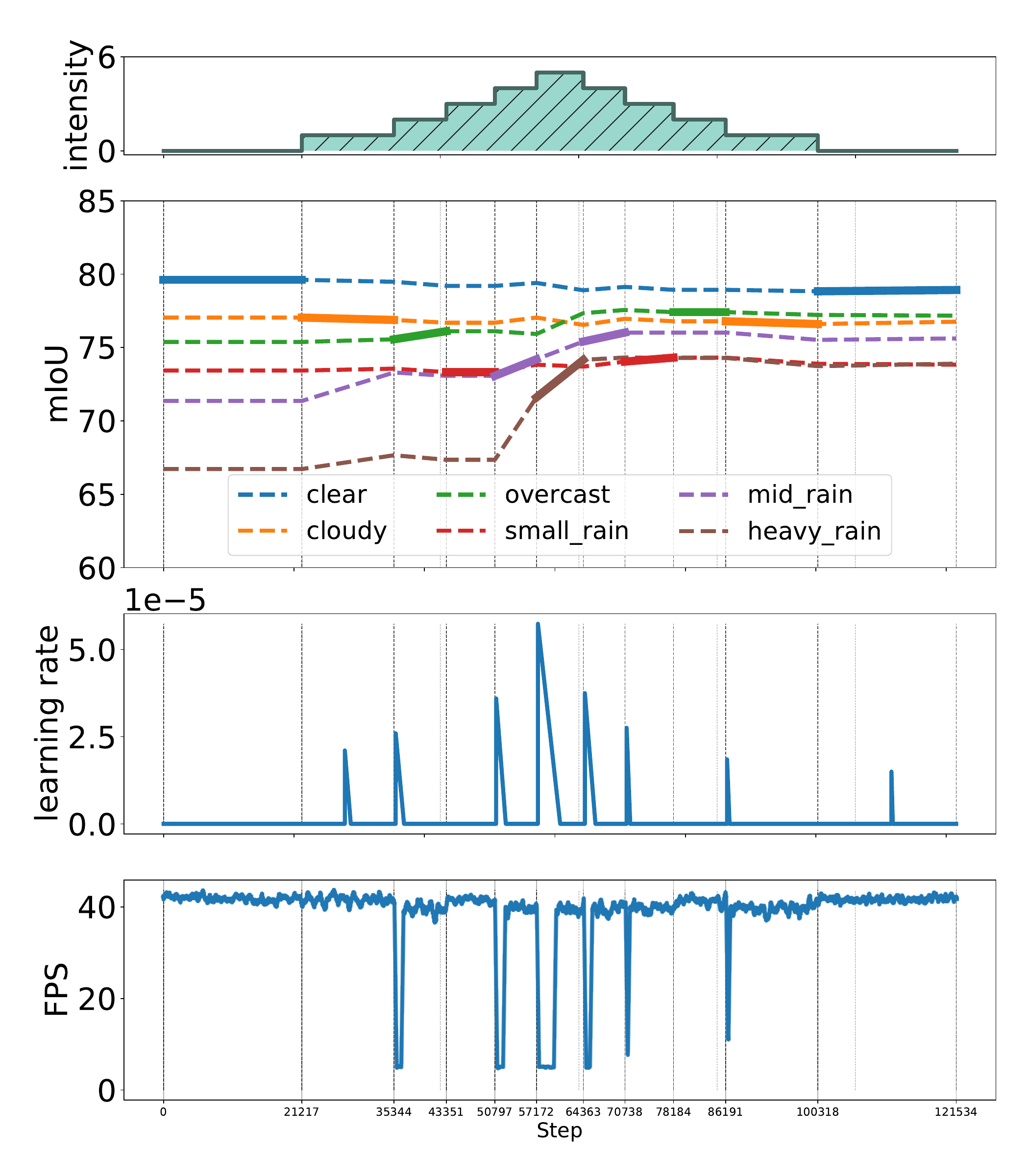}
    \vspace{-0.5cm}
    \caption{\textbf{HAMLET on the SHIFT benchmark.} We show mIoU over active (bold) and inactive (dashed) domains, learning rate and FPS.}
    \label{fig:shift}
\end{figure}

\begin{figure*}[t]
    \centering
    \renewcommand{\tabcolsep}{1pt}
    \begin{tabular}{cccc}
    \textit{clean} & \textit{50mm} & \textit{100mm} & \textit{200mm} \\
    \includegraphics[width=0.21\textwidth]{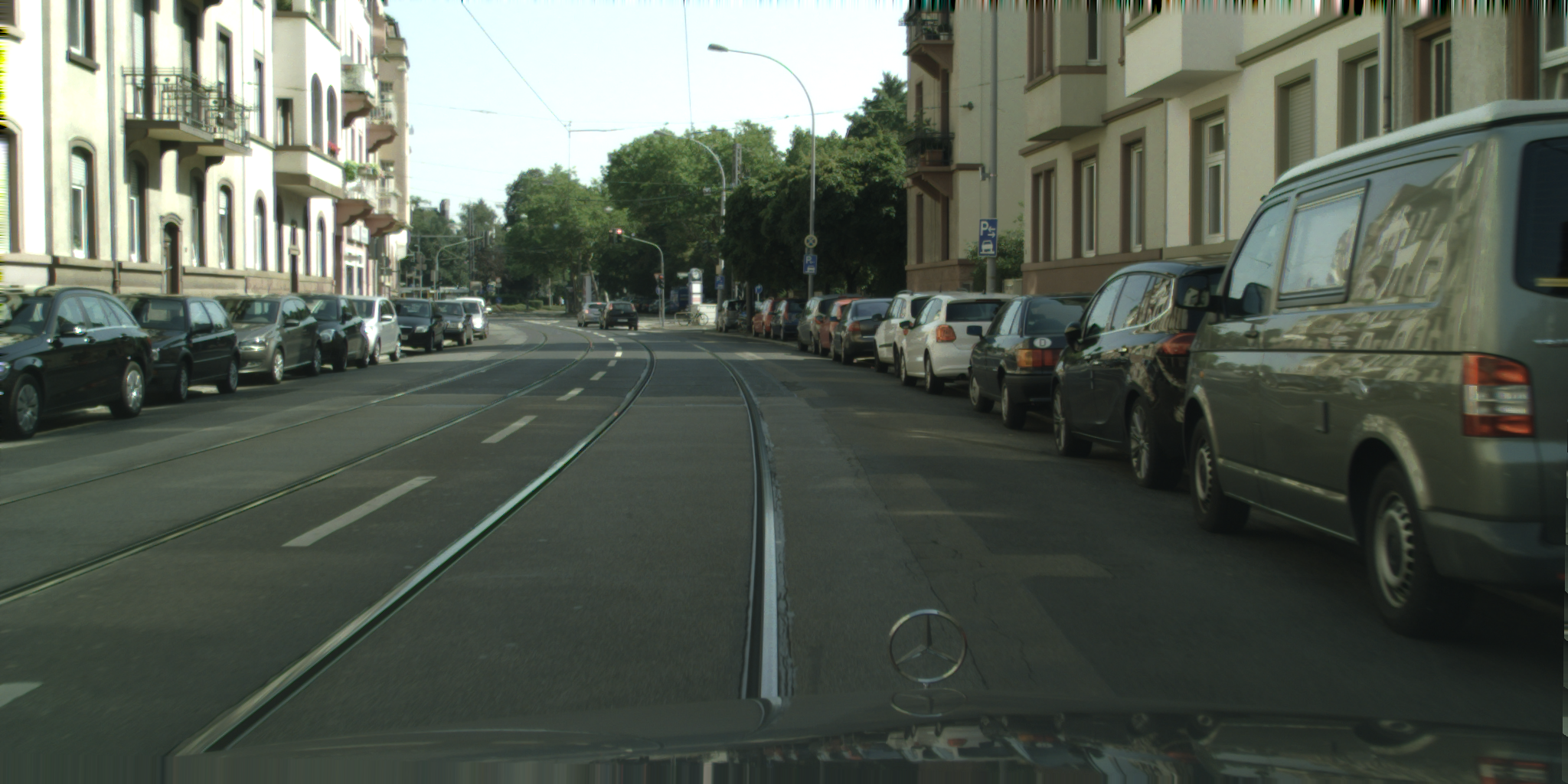} &
    \includegraphics[width=0.21\textwidth]{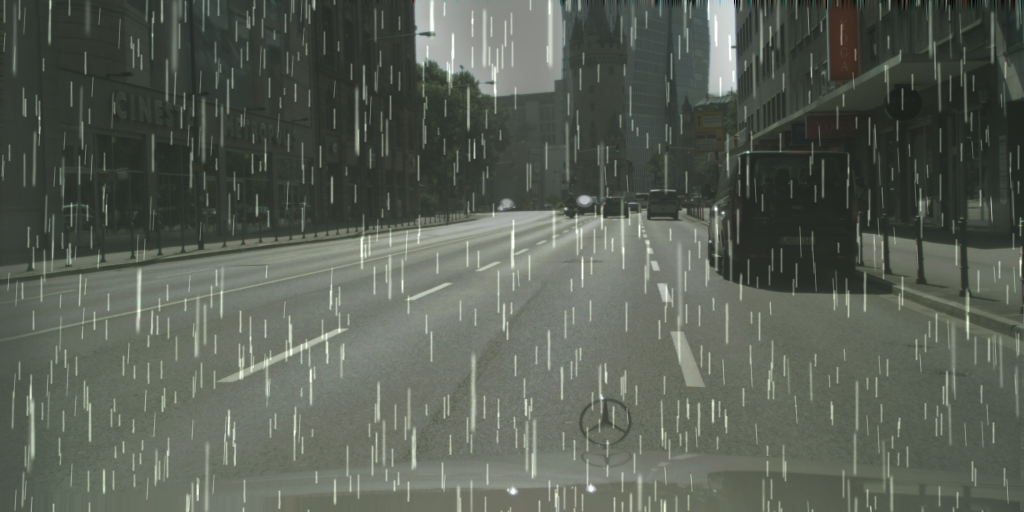} &
    \includegraphics[width=0.21\textwidth]{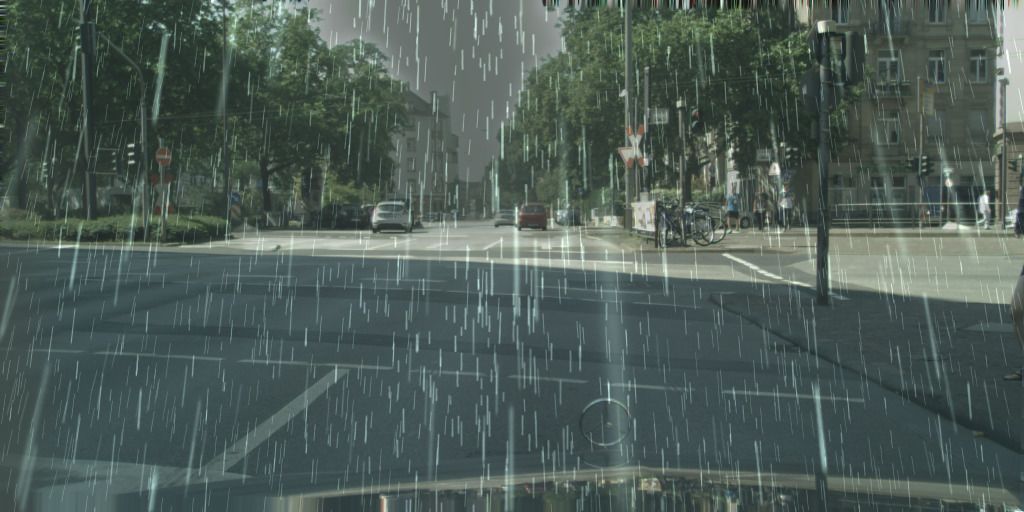} &
    \includegraphics[width=0.21\textwidth]{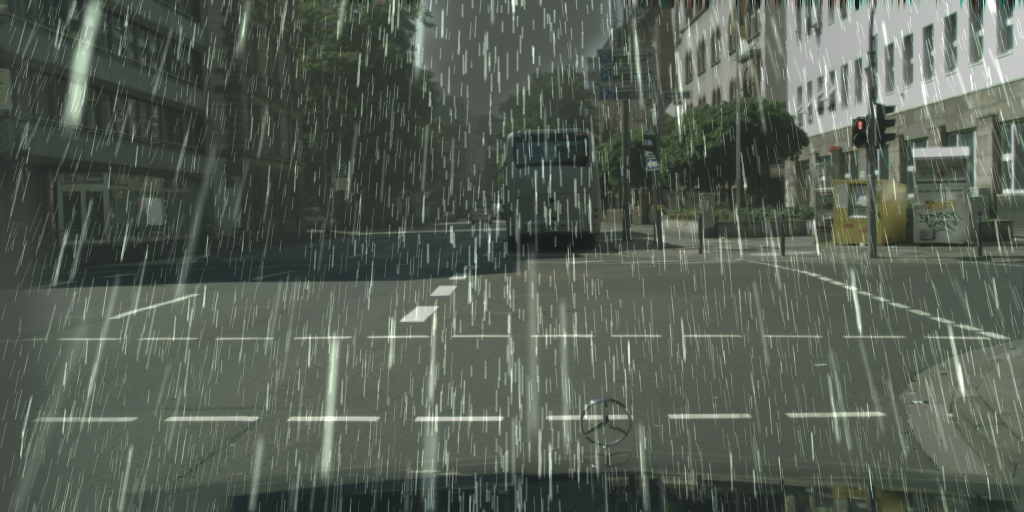} \\
    \includegraphics[width=0.21\textwidth]{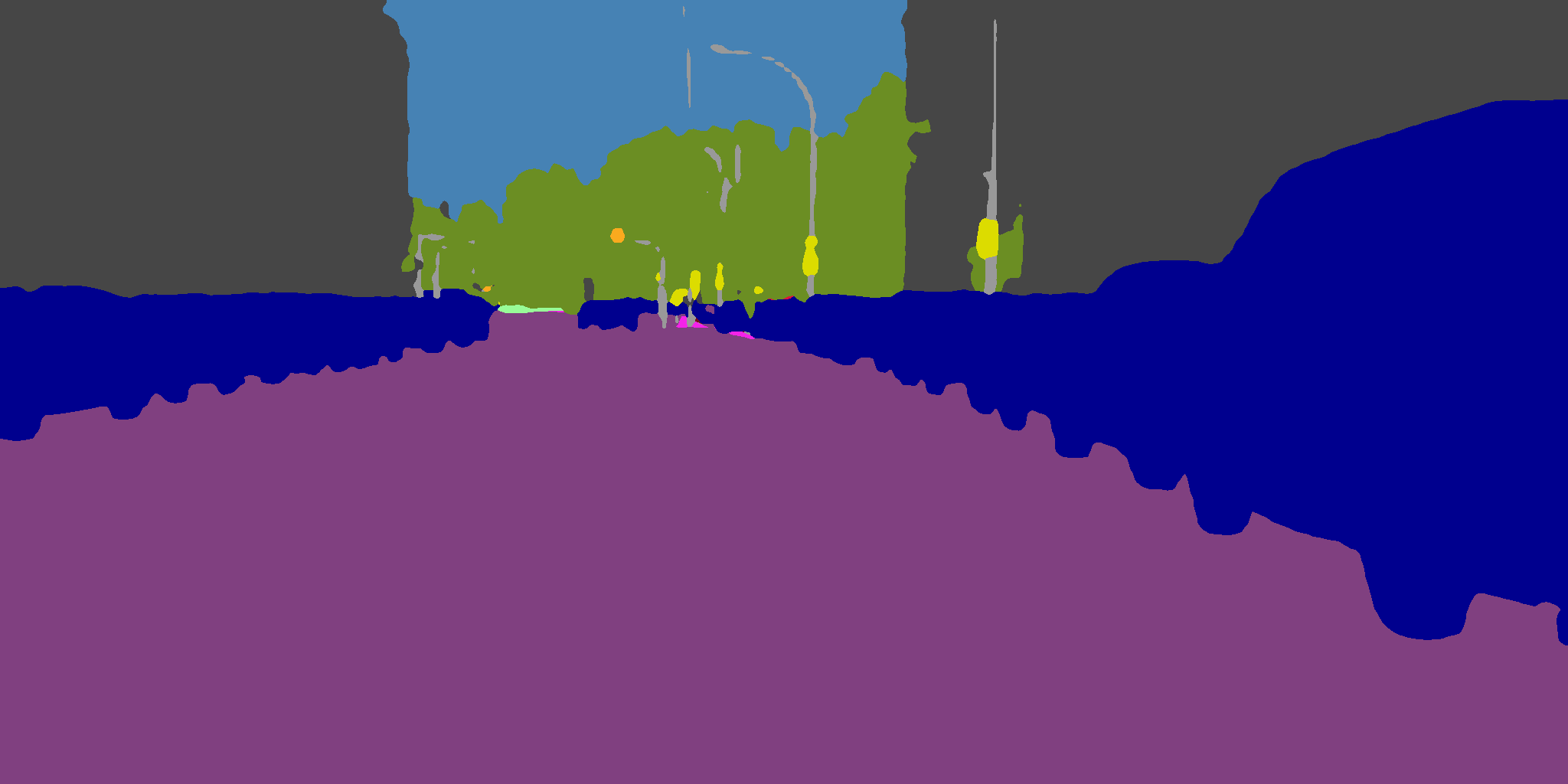} &
    \includegraphics[width=0.21\textwidth]{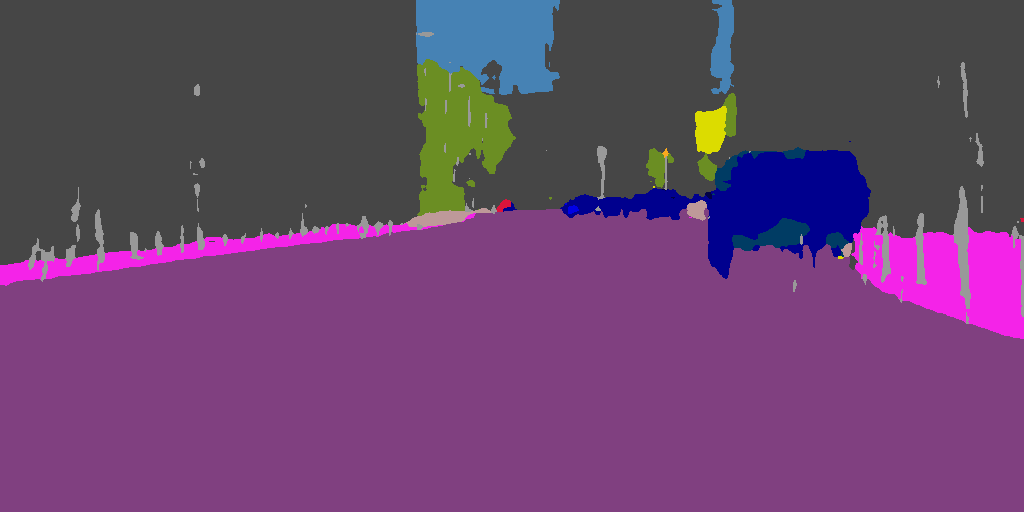} &
    \includegraphics[width=0.21\textwidth]{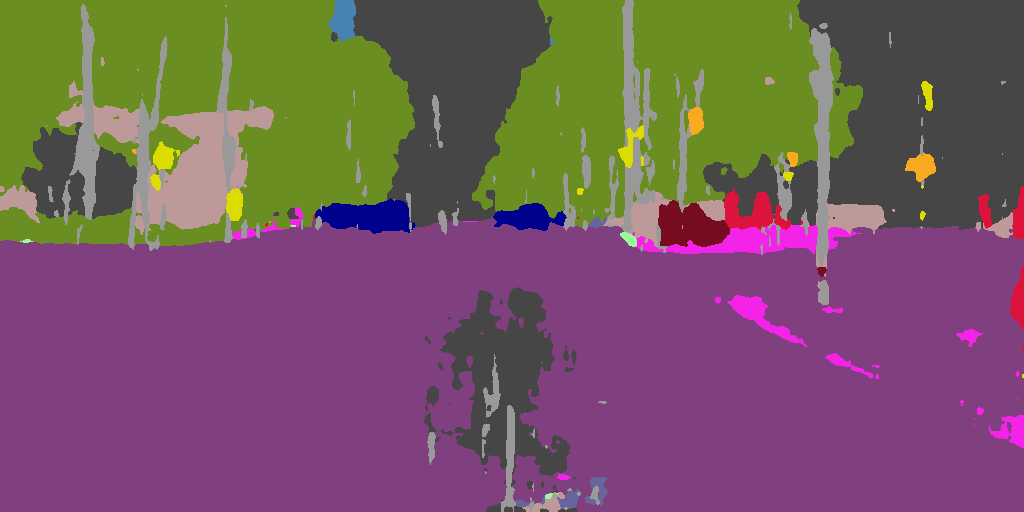} &
    \includegraphics[width=0.21\textwidth]{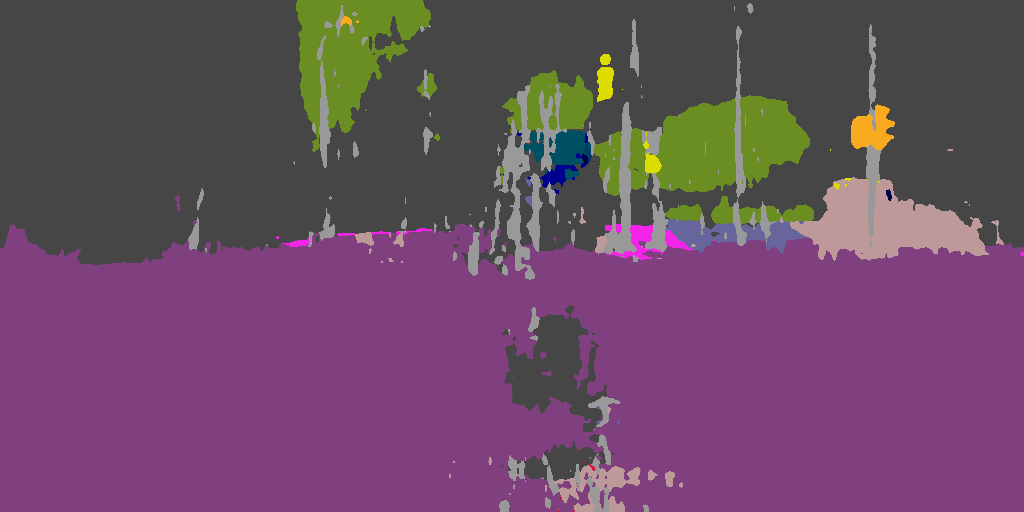} \\
    \includegraphics[width=0.21\textwidth]{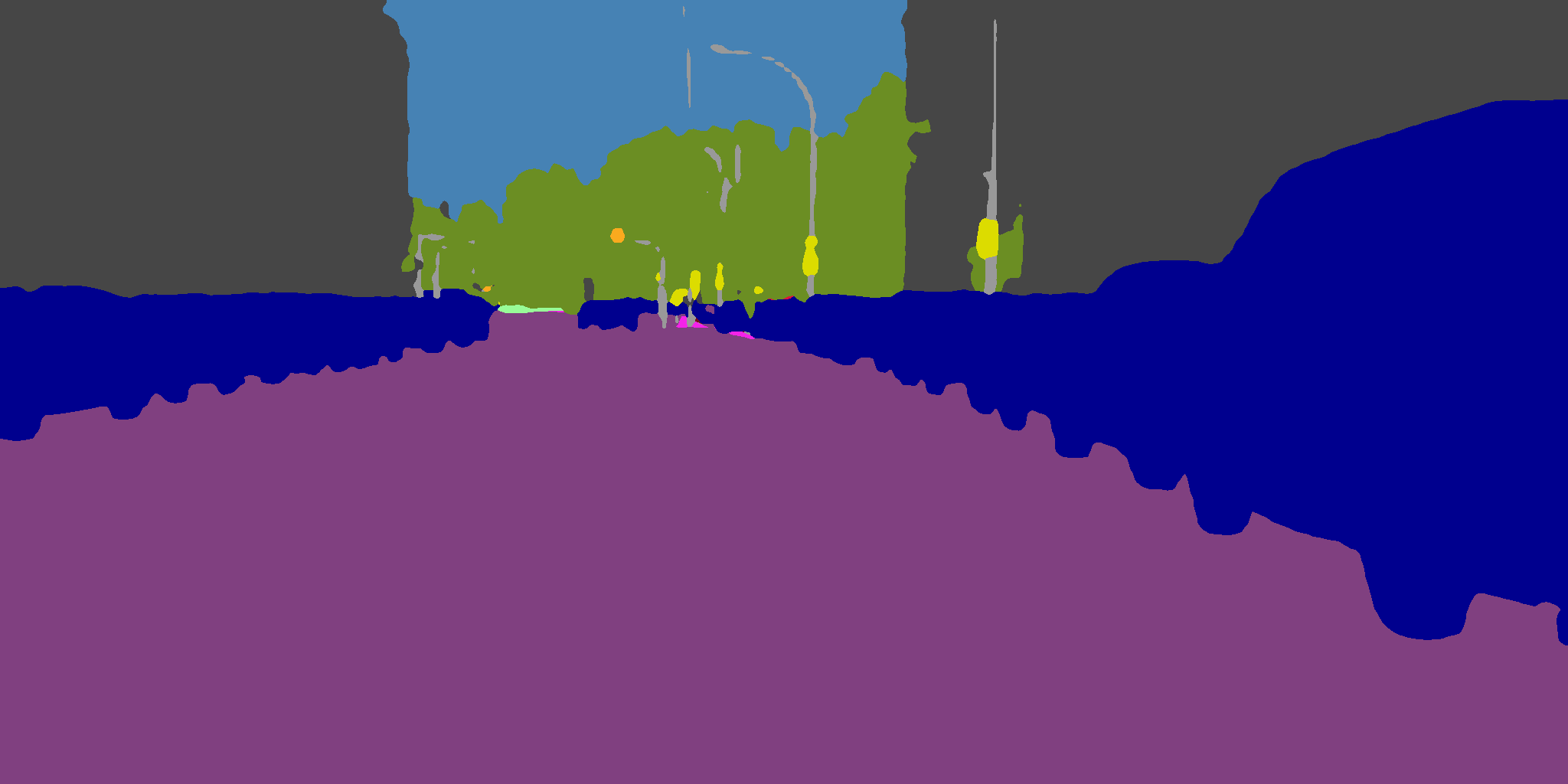} &
    \includegraphics[width=0.21\textwidth]{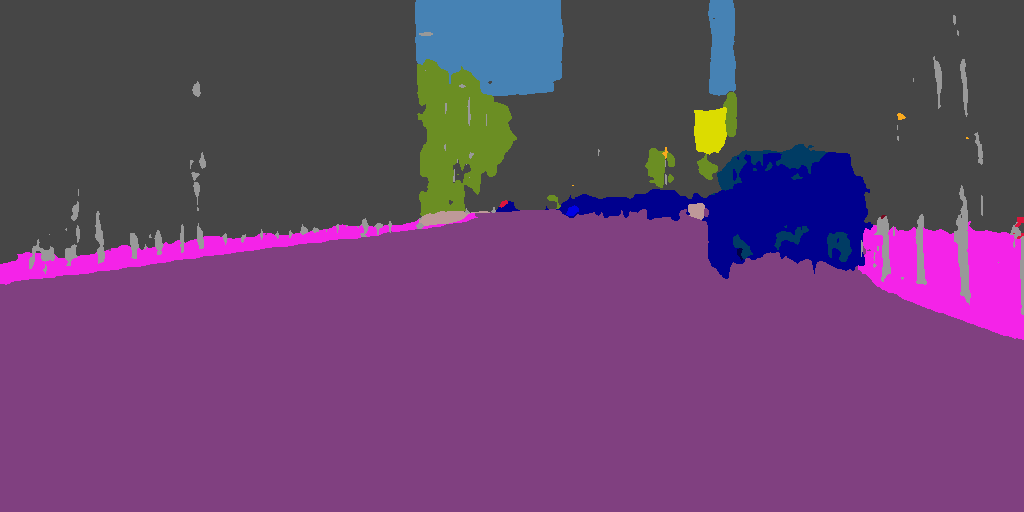} &
    \includegraphics[width=0.21\textwidth]{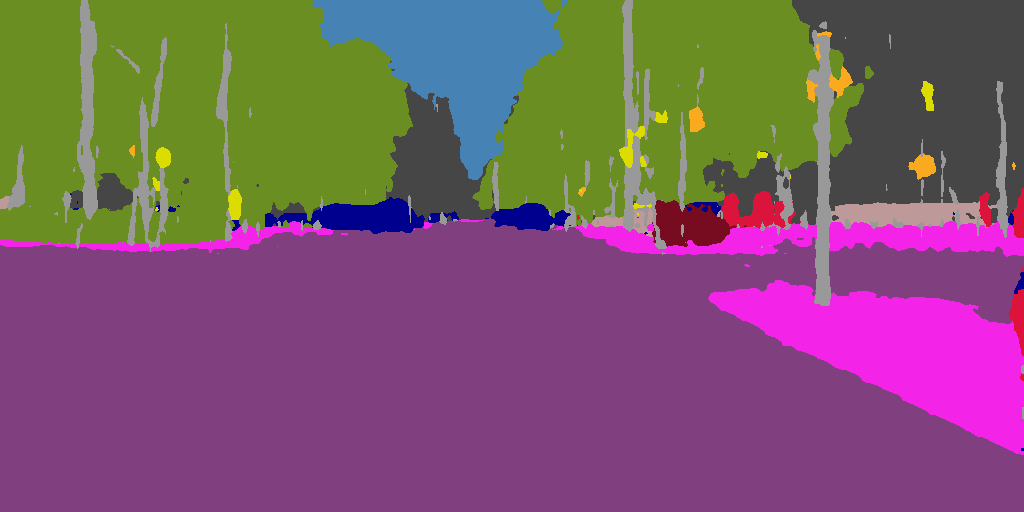} &
    \includegraphics[width=0.21\textwidth]{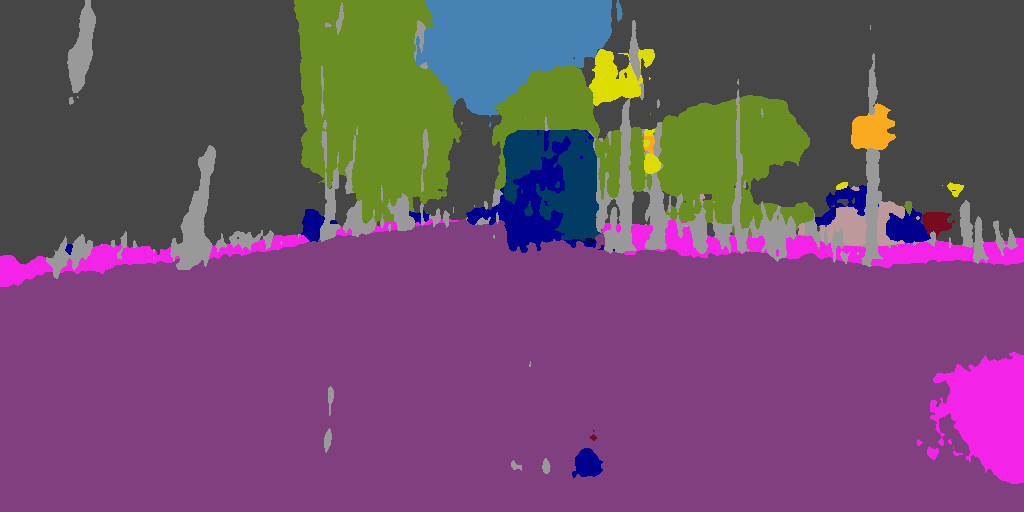} \\
    \end{tabular}\vspace{-0.3cm}
    \caption{\textbf{Qualitative results -- HAMLET in action.} From left to right, we show frames from \textit{clean}, \textit{50mm}, \textit{100mm}, and \textit{200m} domains. From top to bottom: input image, prediction by SegFormer trained on source domain and HAMLET. }
    \label{fig:qual}
\end{figure*}

\subsection{Additional Results: Fog and SHIFT}

\textbf{Fog.} In Tab.~\ref{tab:fog}, we investigate adaptation on the Increasing Fog scenario in the OnDA benchmark~\cite{Panagiotakopoulos_ECCV_2022}. Crucially, for this experiment, we keep the same hyperparameters used for the Increasing Storm, since in both cases the starting SegFormer model is trained on the same source domain. This allows for validating how the proposed setting generalizes at dealing with different kind of domain shifts, beyond those considered in the main experiments. We effectively use Increasing Fog as test set, and compare against SegFormer trained on source (no adaptation) and a model that has been adapted by means of full online training optimization (configuration (A) of Table \ref{tab:ablation}). HAMLET is able to adapt almost as well as the full online training model, with less than a 3 mIoU gap, while enjoying real-time adaptation at nearly $5\times$ the speed using just 40\% of the FLOPS.

\textbf{SHIFT.} We further test HAMLET on the SHIFT dataset~\cite{shift}. Tab.~\ref{tab:shift} collects the results achieved by SegFormer trained on source, full online training and HAMLET respectively, both at forward and backward adaptation across \textit{Clear}, \textit{Cloudy}, \textit{Overcast}, \textit{Small rain}, \textit{Mid rain} and \textit{Heavy rain} domains. Here HAMLET results highly competitive with the full training regime, with only 0.5 drop in average mIoU, while being more than $5\times$ faster.
Fig. \ref{fig:shift} depicts, from top to bottom, the rain intensity characterizing any domain encountered on SHIFT, the mIoU achieved both on current (bold) and inactive (dashed) domains, the learning rate changes based on the domain shift detection, and the framerate achieved at any step. We refer to the supplementary material for a deeper analysis.

\textbf{Qualitative results.} To conclude, Fig. \ref{fig:qual} shows some qualitative examples from CityScapes. We can notice how SegFormer accuracy (second tow) drops with severe rain, whereas HAMLET (third row) is capable of keeping the same segmentation quality across the storm.

\section{Discussion}
\textbf{Orthogonality.} HAMT and LT act independently. Indeed, by strongly constraining the adaptation periods through LT, HAMT has a limited margin of action. The impact of HAMT also depends on the backbone and by carefully crafting modular architectures, one can achieve further optimization. Nevertheless, in a deployment environment where domain shifts occur at high frequencies (\textit{\eg{}}, Storm C), LT is ineffective, while HAMT thrives.

\textbf{Measuring forgetting.} An interesting topic we have not investigated consists of introducing an explicit awareness of which domains have been explored and how well we can recall them, expanding the distance $B$ to multiple dimensions.

\textbf{Safety.} We believe dynamic adaptation has the potential to enhance safety, but we acknowledge the necessity for rigorous testing and verification to safeguard against drift or catastrophic forgetting. This mandates a comprehensive effort from academia, industry, and certification authorities for ensuring the integrity of dynamically adapting models. 

\section{Summary \& Conclusion}
We have presented HAMLET, a framework for real-time adaptation for semantic segmentation that achieves state-of-the-art performance on established benchmarks with continuous domain changes. Our approach combines a hardware-aware backpropagation orchestrator and a specialized domain-shift detector to enable active control over the model's adaptation, resulting in high framerates on a consumer-grade GPU. These advancements enable HAMLET to be a promising solution for in-the-wild deployment, making it a valuable tool for applications that require robust performance in the face of unforeseen domain changes.

\vspace{10pt}
\textbf{Acknowledgement.} The authors thank Gianluca Villani for the insightful discussion on reward-punishment policies, Leonardo Ravaglia for his expertise on hardware-aware training, and Lorenzo Andraghetti for exceptional technical support throughout the project. Their assistance was invaluable in the completion of this work.

{\small
\bibliographystyle{ieee_fullname}
\bibliography{egbib, morebibs}

\begin{thebibliography}{10}\itemsep=-1pt

\bibitem{bobu_adapting_2018}
Andreea Bobu, Judy Hoffman, Eric Tzeng, and Trevor Darrell.
\newblock Adapting to continuously shifting domains.
\newblock In {\em {ICLR} 2018 Workshop Program Chairs}, 2018.
\newblock 00000.

\bibitem{chen2019progressive}
Chaoqi Chen, Weiping Xie, Wenbing Huang, Yu Rong, Xinghao Ding, Yue Huang,
  Tingyang Xu, and Junzhou Huang.
\newblock Progressive feature alignment for unsupervised domain adaptation.
\newblock In {\em Proceedings of the IEEE/CVF Conference on Computer Vision and
  Pattern Recognition}, pages 627--636, 2019.

\bibitem{chen2016semantic}
Liang-Chieh Chen, Jonathan~T Barron, George Papandreou, Kevin Murphy, and
  Alan~L Yuille.
\newblock Semantic image segmentation with task-specific edge detection using
  cnns and a discriminatively trained domain transform.
\newblock In {\em Proceedings of the IEEE conference on computer vision and
  pattern recognition}, pages 4545--4554, 2016.

\bibitem{chen2020naive}
Liang-Chieh Chen, Raphael~Gontijo Lopes, Bowen Cheng, Maxwell~D Collins, Ekin~D
  Cubuk, Barret Zoph, Hartwig Adam, and Jonathon Shlens.
\newblock Naive-student: Leveraging semi-supervised learning in video sequences
  for urban scene segmentation.
\newblock In {\em European Conference on Computer Vision}, pages 695--714.
  Springer, 2020.

\bibitem{chen2017deeplab}
Liang-Chieh Chen, George Papandreou, Iasonas Kokkinos, Kevin Murphy, and Alan~L
  Yuille.
\newblock Deeplab: Semantic image segmentation with deep convolutional nets,
  atrous convolution, and fully connected crfs.
\newblock {\em IEEE transactions on pattern analysis and machine intelligence},
  40(4):834--848, 2017.

\bibitem{deeplabv2}
Liang-Chieh Chen, George Papandreou, Iasonas Kokkinos, Kevin Murphy, and
  Alan~L. Yuille.
\newblock Deeplab: Semantic image segmentation with deep convolutional nets,
  atrous convolution, and fully connected crfs.
\newblock {\em IEEE Transactions on Pattern Analysis and Machine Intelligence},
  40(4):834–848, Apr 2018.

\bibitem{chen2018encoder}
Liang-Chieh Chen, Yukun Zhu, George Papandreou, Florian Schroff, and Hartwig
  Adam.
\newblock Encoder-decoder with atrous separable convolution for semantic image
  segmentation.
\newblock In {\em Proceedings of the European conference on computer vision
  (ECCV)}, pages 801--818, 2018.

\bibitem{chen_no_2017}
Yi-Hsin Chen, Wei-Yu Chen, Yu-Ting Chen, Bo-Cheng Tsai, Yu-Chiang~Frank Wang,
  and Min Sun.
\newblock No more discrimination: Cross city adaptation of road scene
  segmenters.
\newblock In {\em 2017 {IEEE} International Conference on Computer Vision
  ({ICCV})}, pages 2011--2020. {IEEE}, 2017.
\newblock 00000.

\bibitem{cheng2022stochastic}
Feng Cheng, Mingze Xu, Yuanjun Xiong, Hao Chen, Xinyu Li, Wei Li, and Wei Xia.
\newblock Stochastic backpropagation: A memory efficient strategy for training
  video models.
\newblock In {\em Proceedings of the IEEE/CVF Conference on Computer Vision and
  Pattern Recognition}, pages 8301--8310, 2022.

\bibitem{ding2019boundary}
Henghui Ding, Xudong Jiang, Ai~Qun Liu, Nadia~Magnenat Thalmann, and Gang Wang.
\newblock Boundary-aware feature propagation for scene segmentation.
\newblock In {\em Proceedings of the IEEE/CVF International Conference on
  Computer Vision}, pages 6819--6829, 2019.

\bibitem{dosovitskiy2020image}
Alexey Dosovitskiy, Lucas Beyer, Alexander Kolesnikov, Dirk Weissenborn,
  Xiaohua Zhai, Thomas Unterthiner, Mostafa Dehghani, Matthias Minderer, Georg
  Heigold, Sylvain Gelly, et~al.
\newblock An image is worth 16x16 words: Transformers for image recognition at
  scale.
\newblock {\em arXiv preprint arXiv:2010.11929}, 2020.

\bibitem{stylization}
A. {Dundar}, M.~Y. {Liu}, Z. {Yu}, T.~C. {Wang}, J. {Zedlewski}, and J.
  {Kautz}.
\newblock Domain stylization: A fast covariance matching framework towards
  domain adaptation.
\newblock {\em IEEE Transactions on Pattern Analysis and Machine Intelligence},
  pages 1--1, 2020.

\bibitem{fu2019dual}
Jun Fu, Jing Liu, Haijie Tian, Yong Li, Yongjun Bao, Zhiwei Fang, and Hanqing
  Lu.
\newblock Dual attention network for scene segmentation.
\newblock In {\em Proceedings of the IEEE/CVF conference on computer vision and
  pattern recognition}, pages 3146--3154, 2019.

\bibitem{fu2019adaptive}
Jun Fu, Jing Liu, Yuhang Wang, Yong Li, Yongjun Bao, Jinhui Tang, and Hanqing
  Lu.
\newblock Adaptive context network for scene parsing.
\newblock In {\em Proceedings of the IEEE/CVF International Conference on
  Computer Vision}, pages 6748--6757, 2019.

\bibitem{ganin}
Yaroslav Ganin, Evgeniya Ustinova, Hana Ajakan, Pascal Germain, Hugo
  Larochelle, Fran\c{c}ois Laviolette, Mario Marchand, and Victor Lempitsky.
\newblock Domain-adversarial training of neural networks.
\newblock {\em The journal of machine learning research}, 17(1):2096–2030,
  Jan. 2016.

\bibitem{gong2022tacs}
Rui Gong, Martin Danelljan, Dengxin Dai, Danda~Pani Paudel, Ajad Chhatkuli,
  Fisher Yu, and Luc Van~Gool.
\newblock Tacs: Taxonomy adaptive cross-domain semantic segmentation.
\newblock In {\em European Conference on Computer Vision}, pages 19--35.
  Springer, 2022.

\bibitem{he2019adaptive}
Junjun He, Zhongying Deng, Lei Zhou, Yali Wang, and Yu Qiao.
\newblock Adaptive pyramid context network for semantic segmentation.
\newblock In {\em Proceedings of the IEEE/CVF Conference on Computer Vision and
  Pattern Recognition}, pages 7519--7528, 2019.

\bibitem{cycada}
Judy Hoffman, Eric Tzeng, Taesung Park, Jun-Yan Zhu, Phillip Isola, Kate
  Saenko, Alexei Efros, and Trevor Darrell.
\newblock {C}y{CADA}: Cycle-consistent adversarial domain adaptation.
\newblock In Jennifer Dy and Andreas Krause, editors, {\em Proceedings of the
  35th International Conference on Machine Learning}, volume~80 of {\em
  Proceedings of Machine Learning Research}, pages 1989--1998,
  Stockholmsmässan, Stockholm Sweden, 10--15 Jul 2018. PMLR.

\bibitem{hoffman_fcns_2016}
Judy Hoffman, Dequan Wang, Fisher Yu, and Trevor Darrell.
\newblock {FCNs} in the wild: Pixel-level adversarial and constraint-based
  adaptation.
\newblock {\em CoRR}, 2016.
\newblock 00000.

\bibitem{hoyer2021daformer}
Lukas Hoyer, Dengxin Dai, and Luc Van~Gool.
\newblock Daformer: Improving network architectures and training strategies for
  domain-adaptive semantic segmentation.
\newblock {\em arXiv preprint arXiv:2111.14887}, 2021.

\bibitem{hoyer2022hrda}
Lukas Hoyer, Dengxin Dai, and Luc Van~Gool.
\newblock Hrda: Context-aware high-resolution domain-adaptive semantic
  segmentation.
\newblock In {\em European Conference on Computer Vision (ECCV)}, 2022.

\bibitem{iwasawa2021testtime}
Yusuke Iwasawa and Yutaka Matsuo.
\newblock Test-time classifier adjustment module for model-agnostic domain
  generalization.
\newblock In A. Beygelzimer, Y. Dauphin, P. Liang, and J.~Wortman Vaughan,
  editors, {\em Advances in Neural Information Processing Systems}, 2021.

\bibitem{jiang2019accelerating}
Angela~H Jiang, Daniel L-K Wong, Giulio Zhou, David~G Andersen, Jeffrey Dean,
  Gregory~R Ganger, Gauri Joshi, Michael Kaminksy, Michael Kozuch, Zachary~C
  Lipton, et~al.
\newblock Accelerating deep learning by focusing on the biggest losers.
\newblock {\em arXiv preprint arXiv:1910.00762}, 2019.

\bibitem{jiang2022prototypical}
Zhengkai Jiang, Yuxi Li, Ceyuan Yang, Peng Gao, Yabiao Wang, Ying Tai, and
  Chengjie Wang.
\newblock Prototypical contrast adaptation for domain adaptive semantic
  segmentation.
\newblock In {\em European Conference on Computer Vision}, pages 36--54.
  Springer, 2022.

\bibitem{ltir}
Myeongjin Kim and Hyeran Byun.
\newblock Learning texture invariant representation for domain adaptation of
  semantic segmentation.
\newblock {\em 2020 IEEE/CVF Conference on Computer Vision and Pattern
  Recognition (CVPR)}, Jun 2020.

\bibitem{kirkpatrick_overcomming_2016}
James Kirkpatrick, Razvan Pascanu, Neil Rabinowitz, Joel Veness, Guillaume
  Desjardins, Andrei Rusu, Kieran Milan, John Quan, Tiago Ramalho, Agnieszka
  Grabska-Barwinska, Demis Hassabis, Claudia Clopath, Dharshan Kumaran, and
  Raia Hadsell.
\newblock Overcoming catastrophic forgetting in neural networks.
\newblock {\em Proceedings of the National Academy of Sciences}, 114, 12 2016.

\bibitem{kuznietsov2022towards}
Yevhen Kuznietsov, Marc Proesmans, and Luc Van~Gool.
\newblock Towards unsupervised online domain adaptation for semantic
  segmentation.
\newblock In {\em Proceedings of the IEEE/CVF Winter Conference on Applications
  of Computer Vision}, pages 261--271, 2022.

\bibitem{lai2022decouplenet}
Xin Lai, Zhuotao Tian, Xiaogang Xu, Yingcong Chen, Shu Liu, Hengshuang Zhao,
  Liwei Wang, and Jiaya Jia.
\newblock Decouplenet: Decoupled network for domain adaptive semantic
  segmentation.
\newblock In {\em European Conference on Computer Vision}, pages 369--387.
  Springer, 2022.

\bibitem{lao2020continuous}
Qicheng Lao, Xiang Jiang, Mohammad Havaei, and Yoshua Bengio.
\newblock Continuous domain adaptation with variational domain-agnostic feature
  replay.
\newblock {\em arXiv preprint arXiv:2003.04382}, 2020.

\bibitem{lee2022bi}
Geon Lee, Chanho Eom, Wonkyung Lee, Hyekang Park, and Bumsub Ham.
\newblock Bi-directional contrastive learning for domain adaptive semantic
  segmentation.
\newblock In {\em European Conference on Computer Vision}, pages 38--55.
  Springer, 2022.

\bibitem{li2019expectation}
Xia Li, Zhisheng Zhong, Jianlong Wu, Yibo Yang, Zhouchen Lin, and Hong Liu.
\newblock Expectation-maximization attention networks for semantic
  segmentation.
\newblock In {\em Proceedings of the IEEE/CVF International Conference on
  Computer Vision}, pages 9167--9176, 2019.

\bibitem{bdl}
Yunsheng Li, Lu Yuan, and Nuno Vasconcelos.
\newblock Bidirectional learning for domain adaptation of semantic
  segmentation.
\newblock {\em 2019 IEEE/CVF Conference on Computer Vision and Pattern
  Recognition (CVPR)}, Jun 2019.

\bibitem{liang_we_2020}
Jian Liang, Dapeng Hu, and Jiashi Feng.
\newblock Do we really need to access the source data? source hypothesis
  transfer for unsupervised domain adaptation.
\newblock {\em CoRR}, 2020.

\bibitem{liu2021ttt}
Yuejiang Liu, Parth Kothari, Bastien~Germain van Delft, Baptiste Bellot-Gurlet,
  Taylor Mordan, and Alexandre Alahi.
\newblock {TTT}++: When does self-supervised test-time training fail or thrive?
\newblock In A. Beygelzimer, Y. Dauphin, P. Liang, and J.~Wortman Vaughan,
  editors, {\em Advances in Neural Information Processing Systems}, 2021.

\bibitem{liu_source-free_2021}
Yuang Liu, Wei Zhang, and Jun Wang.
\newblock Source-free domain adaptation for semantic segmentation.
\newblock 2021.

\bibitem{fcn}
Jonathan Long, Evan Shelhamer, and Trevor Darrel.
\newblock Fully convolutional networks for semantic segmentation.
\newblock In {\em IEEE Conference on Computer Vision and Pattern Recognition
  (CVPR)}, 2015.

\bibitem{iast}
Ke Mei, Chuang Zhu, Jiaqi Zou, and Shanghang Zhang.
\newblock Instance adaptive self-training for unsupervised domain adaptation.
\newblock {\em Lecture Notes in Computer Science}, page 415–430, 2020.

\bibitem{nekrasovlight}
Vladimir Nekrasov, Chunhua Shen, and Ian Reid.
\newblock Light-weight refinenet for real-time semantic segmentation.
\newblock In {\em British Conference on Computer Vision (BMVC)}, 2018.

\bibitem{olsson2021classmix}
Viktor Olsson, Wilhelm Tranheden, Juliano Pinto, and Lennart Svensson.
\newblock Classmix: Segmentation-based data augmentation for semi-supervised
  learning.
\newblock In {\em Proceedings of the IEEE/CVF Winter Conference on Applications
  of Computer Vision}, pages 1369--1378, 2021.

\bibitem{pan2022ml}
Fei Pan, Sungsu Hur, Seokju Lee, Junsik Kim, and In~So Kweon.
\newblock Ml-bpm: Multi-teacher learning with bidirectional photometric mixing
  for open compound domain adaptation in semantic segmentation.
\newblock In {\em European Conference on Computer Vision}, pages 236--251.
  Springer, 2022.

\bibitem{Panagiotakopoulos_ECCV_2022}
Theodoros Panagiotakopoulos, Pier~Luigi Dovesi, Linus H{\"a}renstam-Nielsen,
  and Matteo Poggi.
\newblock Online domain adaptation for semantic segmentation in ever-changing
  conditions.
\newblock In {\em European Conference on Computer Vision (ECCV)}, 2022.

\bibitem{gta}
Stephan~R Richter, Vibhav Vineet, Stefan Roth, and Vladlen Koltun.
\newblock Playing for data: Ground truth from computer games.
\newblock In {\em European conference on computer vision}, pages 102--118.
  Springer, 2016.

\bibitem{Stan2021UnsupervisedMA}
Serban Stan and Mohammad Rostami.
\newblock Unsupervised model adaptation for continual semantic segmentation.
\newblock In {\em AAAI}, 2021.

\bibitem{su_gradient_2020}
Peng Su, Shixiang Tang, Peng Gao, Di Qiu, Ni Zhao, and Xiaogang Wang.
\newblock Gradient regularized contrastive learning for continual domain
  adaptation.
\newblock 2020.
\newblock 00000.

\bibitem{shift}
Tao Sun, Mattia Segu, Janis Postels, Yuxuan Wang, Luc Van~Gool, Bernt Schiele,
  Federico Tombari, and Fisher Yu.
\newblock {SHIFT:} a synthetic driving dataset for continuous multi-task domain
  adaptation.
\newblock In {\em Computer Vision and Pattern Recognition}, 2022.

\bibitem{takikawa2019gated}
Towaki Takikawa, David Acuna, Varun Jampani, and Sanja Fidler.
\newblock Gated-scnn: Gated shape cnns for semantic segmentation.
\newblock In {\em Proceedings of the IEEE/CVF international conference on
  computer vision}, pages 5229--5238, 2019.

\bibitem{thomas_pandikow_kim_stanley_grieve_2021}
Phillip Thomas, Lars Pandikow, Alex Kim, Michael Stanley, and James Grieve.
\newblock Open synthetic dataset for improving cyclist detection, Nov 2021.

\bibitem{tonioni2019real}
Alessio Tonioni, Fabio Tosi, Matteo Poggi, Stefano Mattoccia, and Luigi~Di
  Stefano.
\newblock Real-time self-adaptive deep stereo.
\newblock In {\em Proceedings of the IEEE/CVF Conference on Computer Vision and
  Pattern Recognition}, pages 195--204, 2019.

\bibitem{dacs}
Wilhelm Tranheden, Viktor Olsson, Juliano Pinto, and Lennart Svensson.
\newblock Dacs: Domain adaptation via cross-domain mixed sampling.
\newblock In {\em Proceedings of the IEEE/CVF Winter Conference on Applications
  of Computer Vision (WACV)}, pages 1379--1389, January 2021.

\bibitem{tremblay2020rain}
Maxime Tremblay, Shirsendu~S. Halder, Raoul de Charette, and Jean-François
  Lalonde.
\newblock Rain rendering for evaluating and improving robustness to bad
  weather.
\newblock {\em International Journal of Computer Vision}, 2020.

\bibitem{adaptsegnet}
Yi-Hsuan Tsai, Wei-Chih Hung, Samuel Schulter, Kihyuk Sohn, Ming-Hsuan Yang,
  and Manmohan Chandraker.
\newblock Learning to adapt structured output space for semantic segmentation.
\newblock {\em 2018 IEEE/CVF Conference on Computer Vision and Pattern
  Recognition}, Jun 2018.

\bibitem{Volpi_2022_CVPR}
Riccardo Volpi, Pau de Jorge, Diane Larlus, and Gabriela Csurka.
\newblock On the road to online adaptation for semantic image segmentation.
\newblock In {\em The IEEE Conference on Computer Vision and Pattern
  Recognition (CVPR)}, June 2022.

\bibitem{vs2023towards}
Vibashan VS, Poojan Oza, and Vishal~M. Patel.
\newblock Towards online domain adaptive object detection.
\newblock {\em 2023 IEEE/CVF Winter Conference on Applications of Computer
  Vision (WACV)}, Jan 2023.

\bibitem{wang2021tent}
Dequan Wang, Evan Shelhamer, Shaoteng Liu, Bruno Olshausen, and Trevor Darrell.
\newblock Tent: Fully test-time adaptation by entropy minimization.
\newblock In {\em International Conference on Learning Representations}, 2021.

\bibitem{fada}
Haoran Wang, Tong Shen, Wei Zhang, Lingyu Duan, and Tao Mei.
\newblock Classes matter: A fine-grained adversarial approach to cross-domain
  semantic segmentation.
\newblock In {\em The European Conference on Computer Vision (ECCV)}, August
  2020.

\bibitem{wang2018haq}
Kuan Wang, Zhijian Liu, Yujun Lin, Ji Lin, and Song Han.
\newblock Haq: Hardware-aware automated quantization with mixed precision,
  2018.

\bibitem{wang2022continual}
Qin Wang, Olga Fink, Luc~Van Gool, and Dengxin Dai.
\newblock Continual test-time domain adaptation, 2022.

\bibitem{wang2018non}
Xiaolong Wang, Ross Girshick, Abhinav Gupta, and Kaiming He.
\newblock Non-local neural networks.
\newblock In {\em Proceedings of the IEEE conference on computer vision and
  pattern recognition}, pages 7794--7803, 2018.

\bibitem{wei2017minimal}
Bingzhen Wei, Xu Sun, Xuancheng Ren, and Jingjing Xu.
\newblock Minimal effort back propagation for convolutional neural networks.
\newblock {\em arXiv preprint arXiv:1709.05804}, 2017.

\bibitem{wu2022d2ada}
Tsung-Han Wu, Yi-Syuan Liou, Shao-Ji Yuan, Hsin-Ying Lee, Tung-I Chen,
  Kuan-Chih Huang, and Winston~H Hsu.
\newblock D2ada: Dynamic density-aware active domain adaptation for semantic
  segmentation.
\newblock In {\em European Conference on Computer Vision (ECCV)}, 2022.

\bibitem{dcan}
Zuxuan Wu, Xintong Han, Yen-Liang Lin, Mustafa~Gökhan Uzunbas, Tom Goldstein,
  Ser~Nam Lim, and Larry~S. Davis.
\newblock Dcan: Dual channel-wise alignment networks for unsupervised scene
  adaptation.
\newblock {\em Lecture Notes in Computer Science}, page 535–552, 2018.

\bibitem{wu_ace_2019}
Zuxuan Wu, Xin Wang, Joseph Gonzalez, Tom Goldstein, and Larry Davis.
\newblock {ACE}: Adapting to changing environments for semantic segmentation.
\newblock In {\em 2019 {IEEE}/{CVF} International Conference on Computer Vision
  ({ICCV})}, pages 2121--2130. {IEEE}, 2019.

\bibitem{wulfmeier_incremental_2018}
Markus Wulfmeier, Alex Bewley, and Ingmar Posner.
\newblock Incremental adversarial domain adaptation for continually changing
  environments.
\newblock 2018.
\newblock 00000.

\bibitem{xie2021segmenting}
Enze Xie, Wenjia Wang, Wenhai Wang, Peize Sun, Hang Xu, Ding Liang, and Ping
  Luo.
\newblock Segmenting transparent objects in the wild with transformer.
\newblock In {\em IJCAI}, 2021.

\bibitem{xie2021segformer}
Enze Xie, Wenhai Wang, Zhiding Yu, Anima Anandkumar, Jose~M Alvarez, and Ping
  Luo.
\newblock Segformer: Simple and efficient design for semantic segmentation with
  transformers.
\newblock {\em Advances in Neural Information Processing Systems},
  34:12077--12090, 2021.

\bibitem{yang2018denseaspp}
Maoke Yang, Kun Yu, Chi Zhang, Zhiwei Li, and Kuiyuan Yang.
\newblock Denseaspp for semantic segmentation in street scenes.
\newblock In {\em Proceedings of the IEEE conference on computer vision and
  pattern recognition}, pages 3684--3692, 2018.

\bibitem{yang_fda_2020}
Yanchao Yang and Stefano Soatto.
\newblock {FDA}: Fourier domain adaptation for semantic segmentation.
\newblock In {\em 2020 {IEEE}/{CVF} Conference on Computer Vision and Pattern
  Recognition ({CVPR})}, pages 4084--4094. {IEEE}, 2020.

\bibitem{yu2018bisenet}
Changqian Yu, Jingbo Wang, Chao Peng, Changxin Gao, Gang Yu, and Nong Sang.
\newblock Bisenet: Bilateral segmentation network for real-time semantic
  segmentation.
\newblock In {\em Proceedings of the European conference on computer vision
  (ECCV)}, pages 325--341, 2018.

\bibitem{yuan2021segmentation}
Yuhui Yuan, Xiaokang Chen, Xilin Chen, and Jingdong Wang.
\newblock Segmentation transformer: Object-contextual representations for
  semantic segmentation.
\newblock In {\em European Conference on Computer Vision (ECCV)}, 2020.

\bibitem{zhang_prototypical_2021}
Pan Zhang, Bo Zhang, Ting Zhang, Dong Chen, Yong Wang, and Fang Wen.
\newblock Prototypical pseudo label denoising and target structure learning for
  domain adaptive semantic segmentation.
\newblock 2021.

\bibitem{zhang_category_2019}
Qiming Zhang, Jing Zhang, Wei Liu, and Dacheng Tao.
\newblock Category anchor-guided unsupervised domain adaptation for semantic
  segmentation.
\newblock {\em Advances in Neural Information Processing Systems}, 2019.

\bibitem{zhao2017pspnet}
Hengshuang Zhao, Jianping Shi, Xiaojuan Qi, Xiaogang Wang, and Jiaya Jia.
\newblock Pyramid scene parsing network.
\newblock In {\em CVPR}, 2017.

\bibitem{zhou2019context}
Yizhou Zhou, Xiaoyan Sun, Zheng-Jun Zha, and Wenjun Zeng.
\newblock Context-reinforced semantic segmentation.
\newblock In {\em Proceedings of the IEEE/CVF Conference on Computer Vision and
  Pattern Recognition}, pages 4046--4055, 2019.

\bibitem{cyclegan}
Jun-Yan Zhu, Taesung Park, Phillip Isola, and Alexei~A. Efros.
\newblock Unpaired image-to-image translation using cycle-consistent
  adversarial networks.
\newblock {\em 2017 IEEE International Conference on Computer Vision (ICCV)},
  Oct 2017.

\bibitem{zou2018unsupervised}
Yang Zou, Zhiding Yu, BVK Kumar, and Jinsong Wang.
\newblock Unsupervised domain adaptation for semantic segmentation via
  class-balanced self-training.
\newblock In {\em Proceedings of the European conference on computer vision
  (ECCV)}, pages 289--305, 2018.

\bibitem{zou2019confidence}
Yang Zou, Zhiding Yu, Xiaofeng Liu, BVK Kumar, and Jinsong Wang.
\newblock Confidence regularized self-training.
\newblock In {\em Proceedings of the IEEE/CVF International Conference on
  Computer Vision}, pages 5982--5991, 2019.

\bibitem{cbst}
Yang Zou, Zhiding Yu, Xiaofeng Liu, B.~V. K.~Vijaya Kumar, and Jinsong Wang.
\newblock Confidence regularized self-training.
\newblock {\em 2019 IEEE/CVF International Conference on Computer Vision
  (ICCV)}, Oct 2019.

\end{thebibliography}
}

\newpage\phantom{Supplementary}
\multido{\i=1+1}{12}{


\section*{Supplementary material}

# To Adapt or Not to Adapt? Real-Time Adaptation for Semantic Segmentation – Supplementary Material

Marc Botet Colomer[* 1,2]  Pier Luigi Dovesi[* 3 †]
Theodoros Panagiotakopoulos[4]  Joao Frederico Carvalho[1]  Linus Härenstam-Nielsen[5,6]
Hossein Azizpour[2]  Hedvig Kjellström[2,3]  Daniel Cremers[5,6,7]  Matteo Poggi[8]

[1]Univrses  [2]KTH  [3]Silo AI  [4]King  [5]Technical University of Munich
[6]Munich Center for Machine Learning  [7]University of Oxford  [8]University of Bologna

https://marcbotet.github.io/hamlet-web/

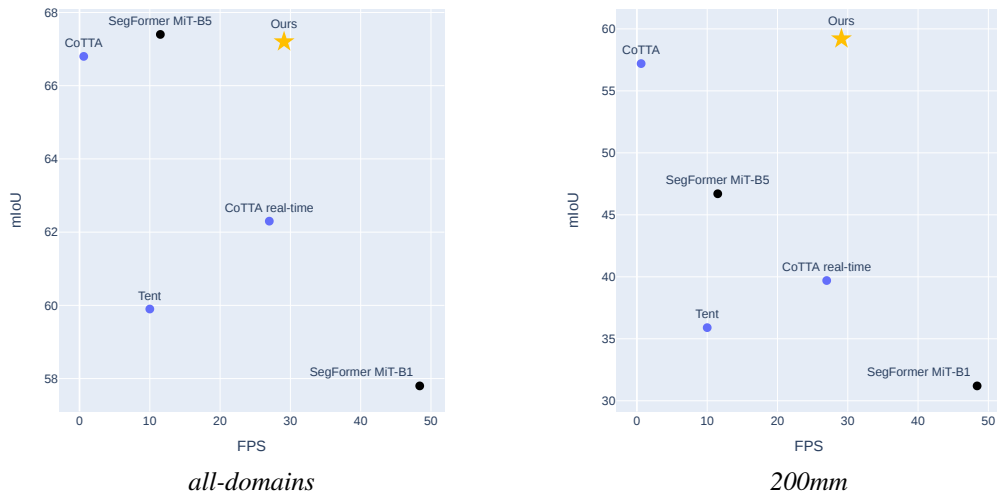

*all-domains*  *200mm*

Figure 1: **Model averaged performances over domain.** We can observe how HAMLET reaches state-of-the-art accuracy, while running more than 6× faster. Adaptive models are displayed in blue, while models trained on source, without adaptation, are displayed in black. On the *left* we show averaged performances over all domains, on the *right* we show how metrics drop in the hardest domain (200mm). The metric drop is limited for strong adaptive networks CoTTA [9], and HAMLET, while being a drastic drop for TENT [8] and SegFormer [10] (both MiT-B1 and MiT-B5). Finally, CoTTA *real-time* shows the performance of CoTTA in deployment conditions, hence running once every 50 frames.

This report introduces further details on the ICCV paper - "To Adapt or Not to Adapt? Real-Time Adaptation for Semantic Segmentation". On the cover of this document, Figure 1, we propose a comparison of HAMLET (Hardware-Aware Modular Least Expensive Training) against state-of-the-art adaptation strategies, hence showing its highly favorable trade-off between speed and accuracy.

Then, starting with Section 1 we present an ablation study on the Hardware-Aware Modular Training (HAMT) method, where we show several speed-accuracy trade-offs. In Section 2 we dive deep in the Least Training (LT) and Adaptive Learning Rate (ALR) methodologies. We illustrate the behavior of different policies by comparing them on the same domain shift, and we also show how the policy copes with noisy domain shift detection. In addition, we present a quantitative analysis focusing on both speed and accuracy. A deeper analysis of the effect of the adaption on every single class is presented in

---

[*] Joint first authorship  [†] Part of the work carried out while at Univrses.

Section 3. Then, in Sections 4 and 5 we detail the model implementation and the chosen hyperparameters. In Section 6 we provide extensive studies on additional storms as presented in the OnDA benchmark [5]. We particularly focus on the behavior of repeated adaptation cycles and how they affect domain shift detection (hence ALR policies). Then, in Section 7 we illustrate additional experiments on the SHIFT dataset [6]. We first present a plot summarizing the model behavior on SHIFT and then we analyze how LT could prevent forgetting in long sequences without relevant domain shifts. We conclude by reporting some qualitative results in Section 8, and by referencing the qualitative videos uploaded on Youtube in Section 9. The first qualitative video (https://www.youtube.com/watch?v=zjxPbCphPDE&t=139s) showcases a comparison between HAMLET, CoTTA, and SegFormer MiT-B1 (no adaptation) on a Cityscapes [2] sequence with the Incremental Storm. Finally, we argue that synthetic data could only partially provide evidence of our methods: purposely, we run HAMLET on a real driving video taken in Korea across different rainy domains – https://www.youtube.com/watch?v=Dwswey-GqQc, whose author gave us consent to use it – to further support its effectiveness. The second qualitative video (https://www.youtube.com/watch?v=zjxPbCphPDE) shows the outcome of this experiment.

# 1. Ablation study: Hardware-Aware Modular Training

In this section, we present an additional ablation study on the HAMT module to further investigate its performance. Table 1 reports the adaptation results achieved by different configurations that exploit HAMT alone, without Active Training Modulation being enabled. We focus on the average adaptation performance and results on the hardest domain (200mm) and source domain (*clear*) to respectively measure the effectiveness of the adaptation scheme on the hardest domain and the robustness when adapting back on the source domain. Additionally, we report the framerate achieved by each configuration and the GFLOPS performed to backpropagate gradients through SegFormer.

We compare the Full training model (A) to several configurations that use a random sampling policy to pick which module to optimize (B, C, D, E) and HAMT sampling strategy (B', C', D', E'), characterized by uniform sampling – not hardware-aware – (B, B') or by setting $\beta$ equal to 2.75, 1.75, or 1 for time conditioning, respectively, for entries (C, C'), (D, D'), and (E, E'). We also include a time-conditioned random sampling policy that uses softmax of the measured FPS for each action, with the temperature controlled by the same parameter $\beta$ introduced in HAMT. The time-conditioned random sampling is achieved by skewing the uniform distribution with a softmax of the measured FPS for each action, effectively making more likely to pick an action, the faster it is. The softmax temperature is controlled by the same parameter $\beta$ introduced in HAMT. This allows us to compare how HAMT performs compared to a simpler baseline bound to achieve similar FPS, nevertheless, the action choice will not be controlled by HAMT reward-punishment algorithm, but it will be randomly sampled.

As expected, we find that the most aggressive GFLOPS reduction corresponds to lower $\beta$ values, but this comes with a price in metrics. Our observations show that even the naïve hardware-aware random policy can significantly reduce the GFLOPS dedicated to backpropagation without drastic metric drops. However, given the same $\beta$, HAMT policy always results in better performance than the naïve time-conditioned policies, both on hard domains (200mm of rain), the source domain (clear, 0mm of rain), and on average (F and T to signal forward and backward adaptation). As reported in the main paper, we set $\beta = 1.75$ as the default choice in any other experiments, allowing for a trade-off between GFLOPS reduction and adaptation effectiveness.

We can notice how the gain lead by HAMT over the naïve the metrics is more prominent for high values of $\beta$ (*i.e.* less intense time conditioning) since it leaves to the reward-punishment algorithm more freedom of action to pick the best modules to train.

Overall, our ablation study shows that HAMT can effectively reduce the computational cost of adaptation without compromising accuracy. HAMT is especially useful when facing harsh and frequent domain shifts, where adaptation cannot be easily interrupted. Moreover, our study provides insights into the impact of time conditioning on the performance of HAMT and the importance of setting appropriate values of $\beta$.

**Focus: Why are we using the $2^{nd}$ derivative and not the $1^{st}$?** Since every action corresponds to an optimization step, we expect that every action will minimize the loss function. Therefore, on average, all actions would receive positive rewards. This might lead to the model repeatedly taking the same action, moreover, we want to reward only those actions which are leading to a sharper loss reduction compared to the other optimization alternatives. Indeed the $2^{nd}$ derivative will be positive only if the loss minimization has been greater than the expected linear extrapolation.

# 2. Ablation study: Active Training Modulation

We now focus on studying variations and single components of the policy we defined in Section 3.3 of the main paper. Specifically, with reference to the notation in Section 3.3, we define 5 different policy variants in an incremental manner, by

|     |                                            | clear |       | ... | 200mm |       | h-mean |       | Backward | % Backward |
|     |                                            | F     | B     |     | F     | F     | B      | T     | GFLOPS   | GFLOPS     |
| --- | ------------------------------------------ | ----- | ----- | --- | ----- | ----- | ------ | ----- | -------- | ---------- |
| (A) | Full training                              | 73.5  | 73.2  | ... | 62.6  | 68.7  | 71.0   | 69.8  | 30.8     | 100.0      |
| (B) | Random policy (uniform)                    | **73.5** | 72.9 | ... | 61.6 | 68.2 | **70.4** | 69.3 | 22.6    | 73.5       |
| (B')| HAMT (no time conditioning)                | **73.5** | **73.1** | ... | **61.9** | **68.4** | **70.4** | **69.4** | **22.4** | **72.9** |
| (C) | Random policy (time-conditioned $\beta = 2.75$) | **73.5** | 73.0 | ... | 59.0 | 67.5 | 70.1 | 68.8 | 21.4 | 69.5 |
| (C')| HAMT $\beta = 2.75$                        | 73.4  | **73.2** | ... | **61.1** | **68.0** | **70.3** | **69.1** | **21.3** | **69.3** |
| (D) | Random policy (time-conditioned $\beta = 1.75$) | 73.3 | 72.8 | ... | 59.9 | 67.6 | **70.4** | 69.0 | 20.7 | 67.3 |
| (D')| HAMT $\beta = 1.75$                        | **73.6** | **72.9** | ... | **60.9** | **67.8** | **70.4** | **69.1** | **20.3** | **65.8** |
| (E) | Random policy (time-conditioned $\beta = 1$) | 73.1 | 72.6 | ... | **60.2** | 67.6 | **70.0** | 68.8 | 19.4 | 63.1 |
| (E')| HAMT $\beta = 1$                           | **73.2** | **72.7** | ... | 60.0 | 67.6 | **70.1** | **68.9** | **17.7** | **57.6** |

Table 1: **Ablation studies – HAMT module (3.2).** We report adaptation results on the Increasing Storm, achieved by exploiting different HAMT configurations. We also report the framerate, as well as the GFLOPS required to perform the backward pass during optimization.

assuming:

**I) Constant learning rate and a number of iterations proportional to $|\Delta B_i|$.** In this policy, it is assumed that the length of the adaptation window should grow with the intensity of the observed domain shift with respect to $z$, the minimum $|\Delta B_i|$ that trigger adaptation. We then compute the number of adaptation iterations as $L = K_l \frac{|\Delta B_i|}{z}$ with the factor $K_l$ and the learning rate $\eta$ kept constant. If new domain shifts are detected before the end of the adaptation windows, the remaining iterations are accumulated.

**II) Constant initial learning rate with decay inversely proportional to $|\Delta B_i|$.** In addition to the previous policy, now the learning rate $\eta$ gradually decays until the adaptation is stopped, the smaller the domain shift, the faster the decay. The initial learning rate $K_\eta$ is kept constant.

**III) Initial learning rate proportional to $|\Delta B_i|$ with constant number of adaptation interations.** This policy assumes to always adapt for a fixed amount of steps, hence $L$ is fixed. However, the initial learning rate is proportional to the intensity of the measured domain shift $|\Delta B_i|$ with respect to the minimum detectable shift $z$. Therefore, the initial learning $K_\eta$, is computed as $K_\eta = P_\eta \frac{|\Delta B_i|}{z}$, where $P_\eta$ is a constant that defines the minimum value of $K_\eta$.

**IV) Number of iterations proportional to $|\Delta B_i|$, and initial learning rate proportional to the discretized distance $B$.** This policy follows II), yet assuming an initial learning rate $K_\eta$ that is higher for domains farther from the source.

$$K_\eta = K_{\eta,\min} + \frac{(B_i - B_{\text{source}})(K_{\eta,\max} - K_{\eta,\min})}{B_{\text{hard}} - B_{\text{source}}} \quad (1)$$

where $B_{\text{source}}$ (resp. $B_{\text{hard}}$) is an estimate of $B$ when the network is close to (resp. far from) the source domain; and $K_{\eta,\min}$ (resp. $K_{\eta,\max}$) is the value of $K_\eta$ assigned when the network is close to (resp. far away from) the source.

**V) Number of iterations proportional to $|S_i|$, direction sensitive, and initial learning rate proportional to the discretized distance $B$.** This is the policy applied in the main paper, building upon policy IV). Here we use a variable multiplicative factor for the number of iterations $K_l$ which depends both on both the distance from the source domain and the direction of the domain shift. The rationale is that domain shifts moving away from the source domain are likely to require a deeper and longer adaptation window. On the contrary, domain shifts moving closer to the source domain require fewer and fewer adaptation steps as we get closer. This is because the model shows good recalling of previously experienced domains as well as presenting strong performances close to the source thanks to regularization strategies we put in place.

$$\tilde{K}_l = \begin{cases} K_{l,\max} & \text{if } S_i \geq 0 \\ K_{l,\min} + \frac{(B_i - B_{\text{source}})(K_{l,\max} - K_{l,\min})}{B_{\text{hard}} - B_{\text{source}}} & \text{otherwise} \end{cases} \quad (2)$$

Where $K_{l,\min}$ (resp. $K_{l,\max}$) is the value of $\tilde{K}_l$ assigned when the network is close to (resp. far away from) the source domain. We will appreciate how this last policy results in the best trade-off between accuracy and speed.

In Tab. 2 we showcase the results achieved on the Increasing Storm by different instances of SegFormer, according to the policy variants outlined so far. In (A) full training is performed, while in (B) and (C) we propose two baselines where we naïvely optimize the model every 15 and 20 frames respectively, or by implementing our policies (I-V). As for HAMT, we report performance on *clear* and 200mm domains, as well as the average forward, backwards and total mIoU together with the framerate. As expected, reducing the adaptation steps to one every 15 or 20 frames definitely increases the FPS,

|  |  |  | clear |  | ... | 200mm | h-mean |  |  | FPS |
| --- | --- | --- | --- | --- | --- | --- | --- | --- | --- | --- |
|  |  |  | F | B |  | F | F | B | T |  |
| (B) | Train every 15 iterations | | 73.2 | 73.3 | ... | 53.3 | 64.1 | 68.3 | 66.2 | 26.5 |
| (C) | Train every 20 iterations | | 73.2 | 73.0 |  | 50.2 | 63.2 | 67.9 | 65.5 | 34.0 |
| (I) | Adapt. iter. (constant $\eta$) | | **73.4** | 72.8 |  | 55.6 | 65.5 | 69.3 | 67.4 | 25.3 |
| (II) | Adapt. iter. | | **73.4** | 73.1 | ... | **58.5** | **66.5** | 69.7 | **68.1** | 25.2 |
| (III) | Adapt. $\eta$ | | **73.4** | 71.4 |  | 55.4 | 65.3 | 67.9 | 66.6 | **31.0** |
| (IV) | Adapt. iter. and $\eta$ | | **73.4** | **73.2** |  | 57.9 | 66.0 | **70.0** | 68.0 | 23.4 |
| (V) | Adapt. iter. and $\eta$, dir. sensitive | | **73.4** | **73.2** |  | 57.8 | 66.0 | 69.0 | 67.5 | 29.1 |

Table 2: **Ablation studies – Active Training Modulation (3.3).** We report adaptation results on the Increasing Storm, achieved by exploiting different Active Training Modulation policies (I-V), together with framerates.

| Domain | Model | Rider | M.bike | Sky | Road | Truck | S.walk | Wall | Veget. | Fence | Tr.Light | Terrain | Bus | Car | Train | Sign | Build. | Person | Pole | Bike | mIoU |
| --- | --- | --- | --- | --- | --- | --- | --- | --- | --- | --- | --- | --- | --- | --- | --- | --- | --- | --- | --- | --- | --- |
| clear | (both) | 53.5 | 61.6 | 94.4 | 98.0 | 77.5 | 83.2 | 56.4 | 92.0 | 53.3 | 62.8 | 63.7 | 78.3 | 93.6 | 56.0 | 72.9 | 91.5 | 76.2 | 57.3 | 72.1 | 73.4 |
| 50mm | No adapt. | 43.8 | 44.5 | 83.5 | 96.3 | 67.1 | 73.2 | 32.5 | 88.0 | 43.4 | 54.0 | 50.6 | 70.5 | 90.6 | 51.1 | 67.5 | 86.3 | 70.1 | 40.4 | 65.7 | 64.2 |
| 50mm | HAMLET | **49.0** | **45.4** | **92.1** | **97.2** | **68.5** | **78.7** | **45.2** | **90.1** | **48.1** | **56.3** | **55.8** | **73.0** | 90.5 | **55.5** | **68.7** | **89.1** | **71.2** | **45.2** | **66.7** | **67.7** |
| 100mm | No adapt. | 30.0 | 24.1 | 37.2 | 92.6 | 54.4 | 57.6 | 19.3 | 80.6 | 30.1 | 41.9 | 40.0 | 58.2 | 85.1 | 45.7 | 60.5 | 75.6 | 63.9 | 28.6 | 58.4 | 51.8 |
| 100mm | HAMLET | **41.6** | **48.3** | **90.6** | **96.8** | **70.0** | **76.1** | **44.5** | **88.7** | **46.7** | **48.3** | **57.9** | **70.7** | **89.5** | **53.1** | **64.3** | **87.5** | **67.7** | **38.3** | **63.5** | **65.5** |
| 200mm | No adapt. | 11.7 | 3.7 | 1.9 | 81.9 | 25.1 | 25.3 | 7.8 | 59.5 | 9.4 | 21.8 | 19.5 | 33.4 | 59.2 | 21.3 | 45.5 | 59.2 | 50.9 | 14.8 | 40.1 | 31.2 |
| 200mm | HAMLET | **36.2** | **32.9** | **85.4** | **96.1** | **65.8** | **71.6** | **33.1** | **86.1** | **40.8** | **38.5** | **53.6** | **69.2** | **86.7** | **35.9** | **57.5** | **84.4** | **62.9** | **29.2** | **59.1** | **59.2** |

Table 3: **Single classes mIoUs.** Results on single classes by the source model on *clear* and *200mm*, and by HAMLET on *50mm*, *100mm*, *200mm* (Incremental storm, forward pass). The improvements achieved with online adaptation are consistent all across the board.

nevertheless it also notably reduces the overall adaptation effectiveness – in particular on the hardest domain of 200mm, with a drop of around 10% compared to full training. To attain a better accuracy-speed trade-off, we employ our policies: we can appreciate how (II) allows for the best overall adaptation performance as well as over 200m of rain while achieving the lowest FPS among the policies. Using (III) we obtain the highest FPS while losing accuracy both on average mIoU and when returning back to the source – specifically, resulting in the worse policy in backward adaptation. Policy (IV) provides for the best backward adaptation results, at the expense of forward adaption and average performance. Finally, (V) balances all of the aspects considered before, while being the second fastest configuration among those considered. We also highlight how these policies merely represent examples of potential uses of the domain detection signals and how even a simple active training configuration policy could enable very fast and effective adaptation processes.

In Figure 2 we provide insight into the Active Training Modulation mechanism. In the top row, we exemplify two simple domain shift sequences: from clear weather to 50mm to clear (on the left) and a much more sudden change, from clear weather to 200mm to 75mm (on the right). In the second row, we display $H$, denoised in $A$ (third row) and discretized in $B$ (fourth row). We then display $S = \Delta B$ (fifth row) acting as the first derivative of $B$ over time frames. On the left, the domain shift is correctly detected as a single domain shift. This is visible by having a single spike in $S$. On the other hand, in the harder scenario (on the right), the domain shift is detected in two consecutive steps. We could define this as a false positive detection. The rows below show how the ALR policies manage the training phases and the learning rate modulation. We remind the reader that when the learning rate is zero, the training phase is inhibited. We notice how the policy formulation can withstand the double-detection of the domain shift by simply recomputing the learning and accumulating the residual iterations, overall presenting a robust behaviour.

## 3. Single Classes Analysis

In Table 3 we present the per-class mIoU of HAMLET and a model just trained on the source domain (No adapt.). We present results on the source domain, where the two models are equivalent, and in the 50mm, 100mm, and 200mm domains of the forward pass of the Incremental Storm [5]. As expected HAMLET vastly improves the non-adapted baseline on each domain, in every single class. Interestingly, we see that HAMLET improvement is not just on a few classes, but instead, all classes are improved by a consistent amount. As expected some classes are more impacted by the domain change, such as the Sky and rare classes (*e.g.* M.bike, Rider, Fence), while some others present greater robustness (*e.g.* Road, Vegetation, Car, Person).

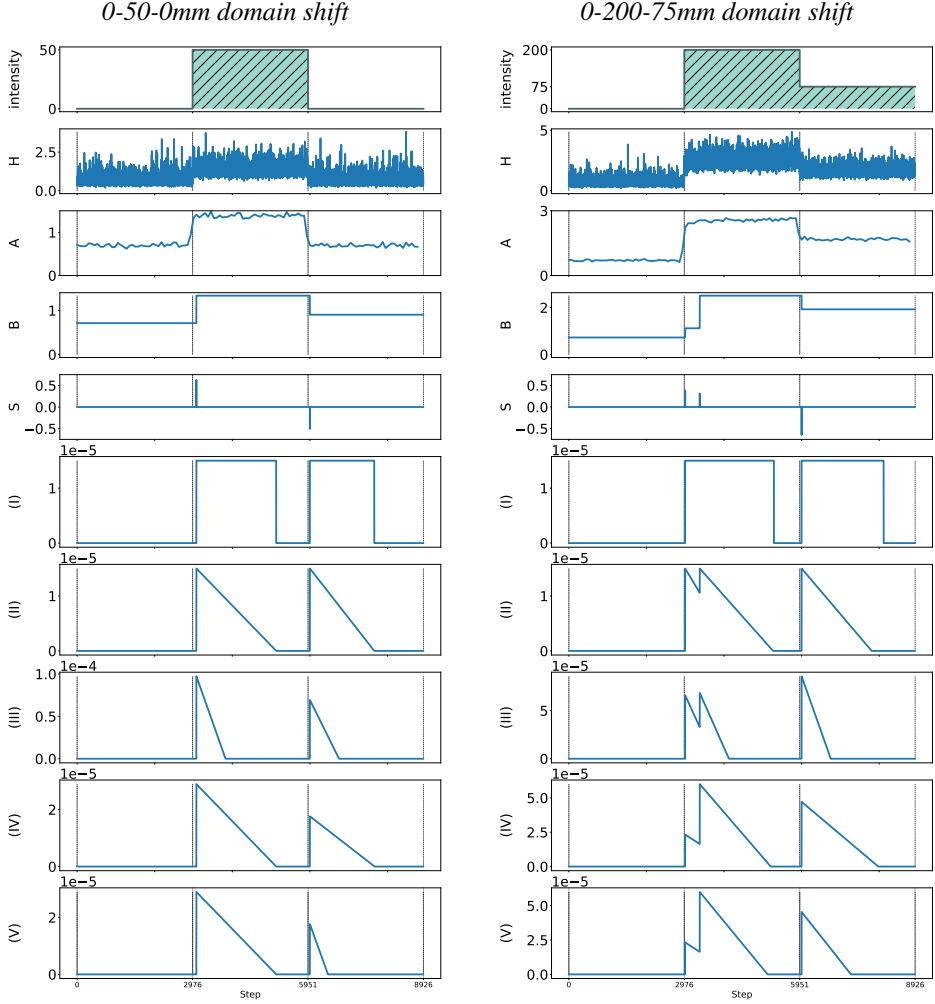

Figure 2: **Focus on the mechanism of domain detection signals and relative adaptation process for each applied policy.**

## 4. Model Setup: Additional Information

In this section, we present additional information on the UDA model and backbone used and provide details about the lightweight decoder.

Let $x \in \mathbb{R}^{C_{in} \times H \times W}$ denote an input image and $y \in [0,1]^{C \times H \times W}$ denote a segmentation label with $C$ number of classes. Let $\mathcal{D}_S = \{(x_S^{(i)}, y_S^{(i)})\}_{i=1}^{n_s}$ be the labeled source dataset and $\mathcal{D}_T = \{x_T^{(i)}\}_{i=1}^{n_t}$ the unlabeled target dataset encounter during deployment, which may contain multiple sequential domains. Our goal is to train a model $f_\theta$ that predicts the probability of each class in each pixel of the input image, such that $f_\theta(x) \in \mathbb{R}^{C \times H \times W}$.

To this end, we use a student-teacher scheme with parameters $\theta$ for the student model and parameters $\theta'$ for the teacher model. During each training iteration $i$, we optimize the student by minimizing the loss function in Eq. 3. The teacher is updated as an EMA of the student weights $f_\theta$ (Eq. 4) where $\alpha$ is the decay rate of the EMA.

$$\mathcal{L}^{(i)} = \mathcal{L}_S^{(i)} + \mathcal{L}_T^{(i)} + \lambda_{FD}\mathcal{L}_{FD}^{(i)} \qquad (3)$$

$$\theta'^{(i+1)} \leftarrow \alpha\theta'^{(i)} + (1-\alpha)\theta^{(i)} \qquad (4)$$

The training loss in Equation 3 is a combination of three terms. The first term, $\mathcal{L}_S$, is a supervised term used to learn the semantic segmentation task using the replay buffer from the labeled dataset and a Cross-Entropy loss. The second term is

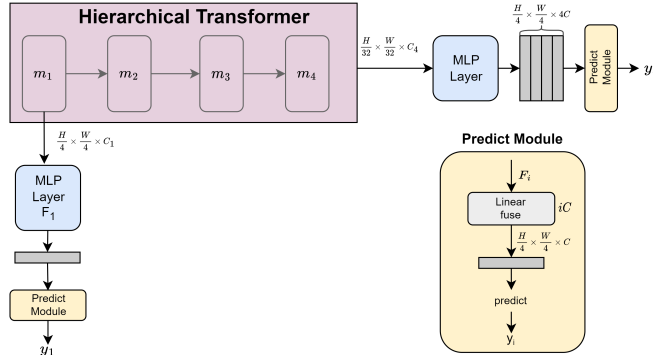

Figure 3: **Adaptive Domain Detector**. We attach a SegFormer all-MLP decoder after the first module $m_1^{\text{fd}}$. This allows us to obtain segmentation maps of any image at a low cost and with very limited speed impact.

a self-training loss that is learned from the target dataset, and the third term is a feature distance loss used as a regularizer. We perform self-training by training the student model $f_\theta$ on a strongly augmented version of the target dataset, along with one-hot encoded pseudo-labels generated by the teacher. The augmented images are generated by mixing randomly selected classes from the source image with target images following ClassMix [4]. $\mathcal{L}_T$ is the cross-entropy between the mixed image and the mixed label weighted by factor $q_T$, as the ratio of the pixels that have a confidence level higher than a certain threshold. To prevent the student network's weights from deviating significantly from a pre-trained model on the source dataset (static teacher), we incorporate a feature distance loss, denoted as $\mathcal{L}FD$, in the training process. Specifically, the feature distance loss is computed by taking the features produced by the student network's encoder $f\theta$ and those produced by a static teacher network with frozen weights on a given input sample, and measuring the Euclidean distance between the feature embeddings generated by these two networks.

For our domain adaptive detector, we utilize SegFormer [10] as a semantic segmentation backbone, incorporating both its encoder and decoder design. We modify the static teacher model $f^{\text{fd}}$ by connecting an extremely lightweight segmentation head, denoted as $d^{\text{fd}}1$, after the first encoder module $m1^{\text{fd}}$, resulting in $h^{\text{fd}}1 = d^{\text{fd}}1 \circ m^{\text{fd}}1$. This lightweight segmentation head follows the SegFormer decoder architecture, using an all-MLP decoder that takes feature encodings from $m^{\text{fd}}1$ with $C_1$ channels and produces segmentation maps using only MLP layers (as illustrated in Fig. 3).

## 5. Implementation details

We report any hyper-parameters used to train the described methods. The supervised models were trained using SegFormer pre-trained weights for 100'000 iterations (selecting the checkpoint with the best validation accuracy) using a learning rate of $6 \times 10^{-5}$, warm-up and linear decay scheduling. The online models were trained using AdamW with $\beta_1 = 0.9, \beta_2 = 0.999$ and weight decay 0.01. The hyperparameters values described in the method section are: $\alpha = 0.1$, $K_l = 750$, $K_{\eta,\min} = 1.5 \times 10^{-4}$ $K_{\eta,\max} = 6 \times 10^{-5}$, $K_{l,\min} = 187$, $K_{l,\max} = 562$, $K_{\text{CM},\min} = 0.5$, $K_{\text{CM},\max} = 0.75$ and $m = 75$. For the storm and fog scenarios we use: $B_{\text{source}} = 0.8$, $B_{\text{hard}} = 2.55$. For the SHIFT dataset [6], we use $B_{\text{source}} = 0.46$, $B_{\text{hard}} = 1.85$ and $m = 200$. For the video sequences, we use the fog and storm parameters with $m = 350$. We also used 1000 source images as a buffer. All models performed training with a batch size of 1 and images scaled to $512 \times 1024$ resolution and random crops of $512 \times 512$ where using for training. Both in HAMLET and in the full training baseline we employ SegFormer decoder, without using DAFormer [3] custom head. It's also worth noting that, to marginalize the impact of different backbones, all tested models in this work are using SegFormer MiT-b1 backbone (*i.e.* HAMLET, TENT, CoTTA) as model backbone, except if specified otherwise (*i.e.* OnDA and Advent [7] are using DeepLabV2 [1]).

During training (evaluation is included), HAMLET consists of the following forward passes:

- Student model using source buffer image
- Static teacher encoder using source buffer image (no decoder)
- EMA teacher using a target image
- Student model using mixed image
- Student model using a target image to provide a prediction
- First module of static teacher in the target image and relative small decoder

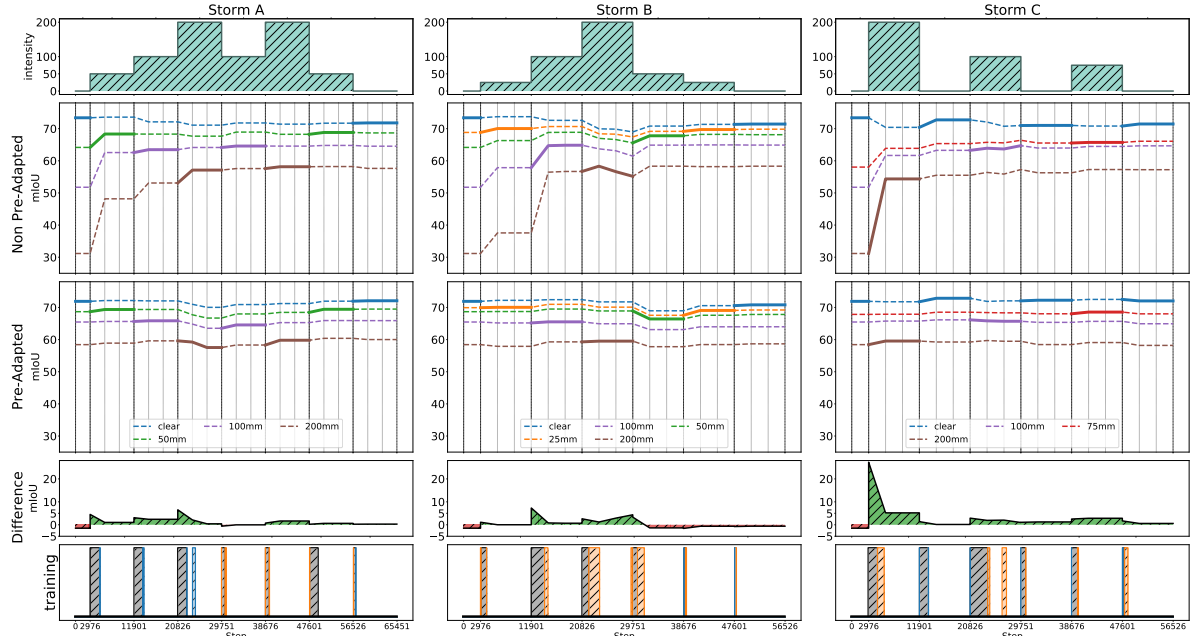

Figure 4: **Experimental results – Storms A, B, C.** Adaptation performance by two HAMLET instances, one trained on source domain (*clear*) and adapted for the first time, and one that has been pre-adapted on the Increasing Storm. In the last two rows, we show the boost in accuracy achieved by the latter model compared to the former, as well as the iterations during which the two are optimized (orange: the former only, blue: the latter only, gray: both).

| | clear | | 50mm | | 10mm | | 200mm | | h-mean | | FPS | GFLOPS |
|---|---|---|---|---|---|---|---|---|---|---|---|---|
| | F | B | F | B | F | B | F | B | F | B | T | |
| Full training | 73.6 | 73.0 | 69.3 | 70.1 | 66.6 | 66.4 | 61.4 | 62.8 | 67.4 | 67.9 | 67.6 | 4.6 | 125.2 |
| HAMLET (non pre-adapted) | **73.4** | 71.8 | 68.3 | 68.8 | 63.5 | **64.6** | 57.1 | 58.2 | 65.0 | 65.4 | 65.2 | **22.8** | **58.2** |
| HAMLET (pre-adapted) | 71.9 | **72.1** | **69.4** | **69.4** | **65.8** | 64.6 | **58.1** | **59.8** | **65.9** | **66.1** | **66.0** | 20.2 | 59.8 |

Table 4: **Experimental results – Storm A**

Backpropagation is applied on the student model only. Afterwards, the dynamic teacher is updated as EMA of the student.

During simple evaluation, HAMLET consists of the following forward passes:

- Student model using a target image to provide a prediction
- First module of static teacher in the target image and relative small decoder

The full source code used for our experiments is attached to this document (**hamlet_code.zip**).

## 6. More storms and longer adaptation analysis

We run HAMLET on three additional rainy scenarios, generated as detailed in [5]. We both evaluate the performance of the brand-new adaptation cycle, starting from SegFormer trained on source domain and adapting to the new storms A, B and C. Additionally, we test a model previously online adapted on the Increasing Storm scenario and compare the two (Non Pre-Adapted and Pre-Adapted) in terms of performance and training phases.

Figure 4 collects, from left to right, the results achieved on Storms A, B, and C as defined in [5]. On top, we plot the rain intensity over time faced during the adaptation process, followed by mIoU plots highlighting how the two models introduced before adapt and the difference in terms of mIoU achieved by the pre-adapted model compared to the brand-new one. On the last row, we show the iterations during which the models are optimized, specifically in gray when both run back-propagation, while in orange and blue when only the brand-new or the pre-adapted model are optimized, respectively.

We notice, similarly to OnDA [5], how HAMLET also benefits from previous adaptation on the Increasing Storm. The highest gain is achieved on Storm C, in which the domain rapidly switches from source to the hardest one, *i.e.* $200mm$. Moreover, we can appreciate in general how the pre-adapted model witnessed almost no drop in accuracy on the inactive domains. This is caused by the Active Training Modulation strategy, which limits the amount of adaptation steps performed

|  | clear | 25mm | 100mm | 200mm | 50mm | 25mm 2 | clear 2 | total h-mean | FPS | GFLOPS |
|---|---|---|---|---|---|---|---|---|---|---|
| Full training | 73.3 | 70.6 | 68.2 | 64.1 | 66.1 | 70.9 | 72.1 | 69.2 | 4.6 | 125.2 |
| HAMLET (non pre-adapted) | **73.4** | **70.0** | 67.4 | 61.6 | 61.5 | **68.8** | **70.6** | 67.3 | 20.0 | 50.3 |
| HAMLET (pre-adapted) | 71.9 | **70.0** | **67.8** | **62.5** | **63.0** | 67.7 | 69.9 | **67.4** | **25.2** | **44.9** |

Table 5: **Experimental results – Storm B**

|  | clear 1 | 200mm | clear 2 | 100mm | clear 3 | 75mm | clear 4 | clear h-mean | target h-mean | total h-mean | FPS | GFLOPS |
|---|---|---|---|---|---|---|---|---|---|---|---|---|
| Full training | 73.6 | 60.1 | 73.4 | 65.6 | 73.0 | 68.8 | 72.9 | 73.2 | 64.6 | 69.3 | 4.5 | 125.2 |
| HAMLET (non pre-adapted) | **73.4** | 54.4 | 72.7 | 64.7 | 71.0 | 65.7 | 71.4 | 72.1 | 61.1 | 67.0 | 16.4 | 51.9 |
| HAMLET (pre-adapted) | 71.9 | **59.5** | **72.9** | **65.7** | **72.2** | **68.6** | **72.0** | **72.3** | **64.4** | **68.7** | **22.2** | **48.3** |

Table 6: **Experimental results – Storm C**

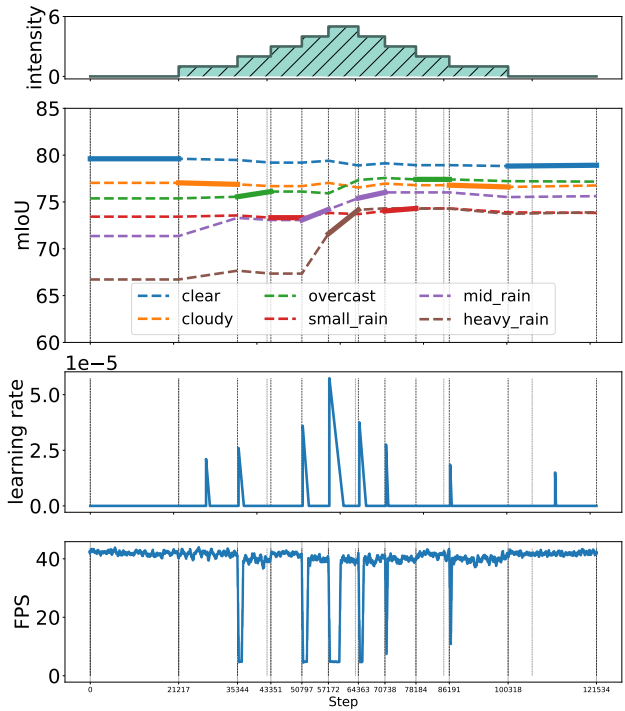

Figure 5: **Experimental results – SHIFT benchmark.** We show mIoU over active (bold) and inactive (dashed) domains, learning rate and FPS.

by HAMLET to be as few as needed to adapt to the new domains while neglecting the occurrence of catastrophic forgetting over inactive domains. Indeed, once adaptation has been performed over the Increasing Storm, it results sufficient to maintain high accuracy when moving to the new storms, as pointed out by the almost horizontal dashed lines in the pre-adapted plots. By focusing on the last row, we point out how, in most times, the non pre-adapted model runs significantly more optimization steps (orange) compared to the pre-adapted one (blue), which can already deal with the domain switches occurring in these storms. This is due to its prior experience on the Increasing Storm, indeed the domain detection relies both on the static lightweight decoder and the student itself. When the student becomes more robust to new domains, also the domain detection becomes more accurate. We also notice how, despite training only a fraction of the iterations, the Non Pre-adapted model can still catch-up with the pre-adapted one, with a delay, even in the hardest transition, *i.e.* storm C. For a quantitative overview of HAMLET performance on the three storms, we collect the results in Tables 4, 5 and 6, respectively for Storms A, B, and C. In particular, we point out how the pre-adapted model is more accurate, as well as faster than the non pre-adapted one on B and C, since it activates adaptation fewer times as previously discussed with reference to Figure 4.

To conclude, HAMLET can benefit from previous adaptation both in terms of accuracy, as well as speed (the fewer the optimization steps, the higher the framerate).

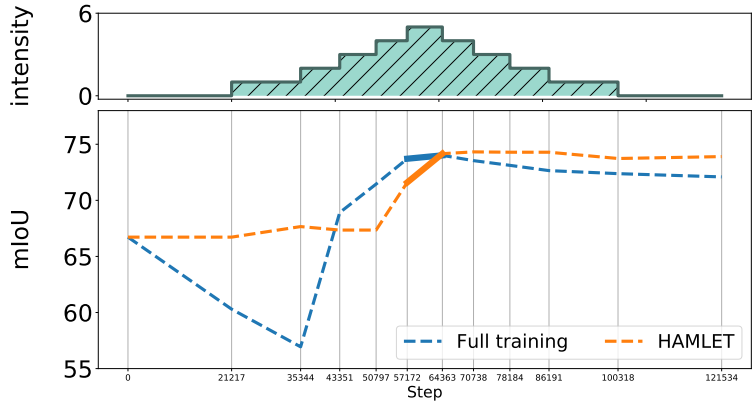

Figure 6: **Experimental results – SHIFT benchmark.** We show the mIoU of the *heavy_rain* domain during the training cycle for Full training and HAMLET. Bold lines represent when the domain is active and dashed lines when is inactive.

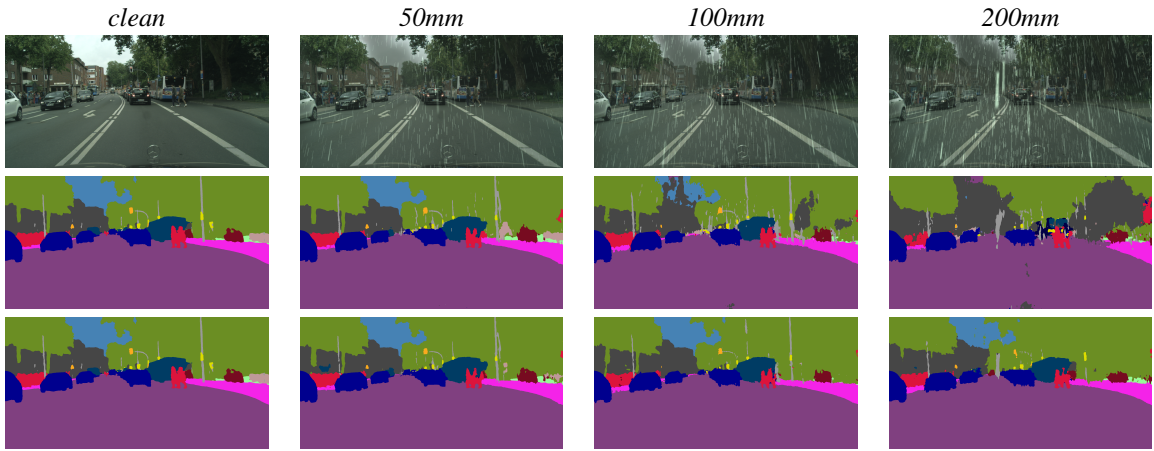

Figure 7: **Qualitative results – HAMLET in action.** From left to right, we show the same source frame, augmented with increasing rain intensity, respectively *clean*, *50mm*, *100mm* and *200m*. From top to bottom: input image, prediction by SegFormer trained on source and HAMLET.

## 7. SHIFT analysis

We now dive deeper in HAMLET performance on the SHIFT benchmark. Figure 5 depicts, on top, the rain intensity characterizing any domain encountered while running HAMLET on SHIFT. Then, we plot the mIoU achieved both on current (bold) and inactive (dashed) domains, as done for the Increasing Storm in the main paper and Storms A, B, C in the previous section. Then, we show how the learning rate changes based on the domain shift detection, followed by the framerate achieved by HAMLET at any step.

From the mIoU plot, we can notice how the drop in accuracy, even on the hardest domain, is moderate compared to what was observed in OnDA benchmarks [5]. We speculate that this might be caused by the full-synthetic nature of this domain, which makes the task easier. Interestingly, the performance on *small_rain* and *mid_rain* domains continue to improve even after HAMLET moves to further domains. In general, as previously observed on Storms A, B, C, HAMLET do not experience any catastrophic forgetting on the inactive domains that have been faced previously. For what concerns domain shift detection, we can notice how this sometimes occurs with a slight delay – *i.e. clear* to *cloudy* and vice-versa – or does not occur at all – *i.e. overcast* to *small_rain* and vice-versa. Nevertheless, once again, this confirms that just a few adaptation steps aligned with the domain shifts are enough to achieve an accuracy comparable to the one obtained with full training, as shown in Tab. 4 in the main paper.

This dataset evaluation offers further insights when it comes to observing another problematic behavior of naïve full

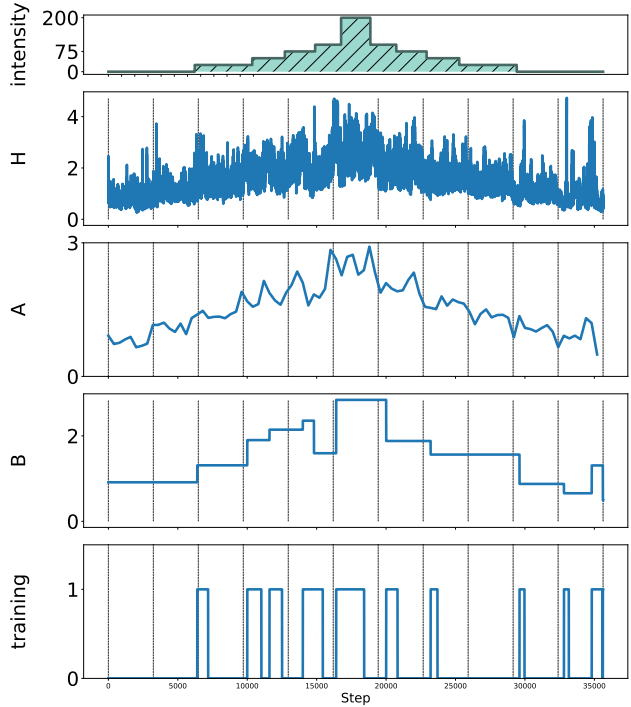

Figure 8: **HAMLET adaptation schedule over the qualitative video sequence.** From first to last row we present: rain intensity, domain detection signals: $H, A, B$, training phases (1: training + inference, 0: inference)

training. Indeed, besides being vastly more computationally expensive, training when it is not required, contributes to the futile specialization of specific domains, hence leading to worse generalization on other domains. This is clearly visible in Figure 6 when we focus on the performances of our Full Training baseline on the *heavy_rain* domain. During the adaptation to clear weather, we notice how evaluating on *heavy_rain* leads to progressively worse performances without achieving any significant improvement in clear weather either. Despite its ability to eventually adapt, this behavior might raise concerns when it comes to sudden domain shifts and it hints to potential domain forgetting. To support this, we show in Figure 6 the accuracy achieved on the *heavy_rain* domain at any time during adaptation, when being the active (bold) or inactive (dashed) domain, for both SegFormer adapted with full training and HAMLET. We can notice how the full training regime leads to dramatic drops in accuracy on this domain when it is inactive, until it is actually encountered. HAMLET, on the contrary, can preserve its original accuracy on the *heavy_rain*, proving that selective adaptation also avoids catastrophic forgetting, to which full training is prone to.

## 8. Qualitatives

In Figure 7 we present extensive qualitative examples from the Increasing Storm evaluation set. Figure 7 shows the results achieved by the source model and HAMLET on increasing rain intensity. We can notice how the source model, at first, is robust to moderate rain. When moving towards higher intensity, the model gradually starts failing, whereas HAMLET keeps high accuracy.

## 9. Videos

To conclude, we attach two qualitative videos to this document. For the first (https://www.youtube.com/watch?v=zjxPbCphPDE&t=139s) we emulate a realistic deployment by synthesizing rain over Cityscapes. The domain shift sequence follows the same pattern as the Increasing Storm. In this case, we cap the video framerate at 5.88FPS using the same setup of [5]. On this video, we run SegFormer in three main flavors: 1) trained on source domain, 2) adapting using CoTTA and 3) adapting with HAMLET. We mainly compare against CoTTA: while HAMLET keeps the pace with the considered framerate, CoTTA – which runs at 0.6FPS – is trained over 1 frame every 10, allowing it to keep the pace with the

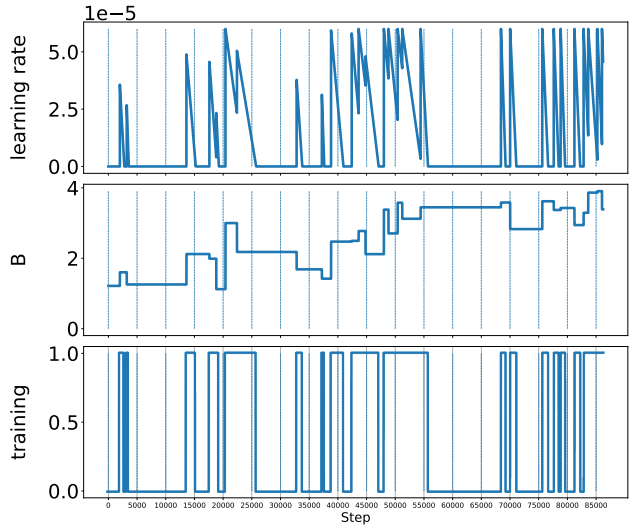

Figure 9: **HAMLET adaptation schedule over the qualitative video sequence.** From first to last row we present: learning rate schedule, domain detection signal $B$, training phases (1: training + inference, 0: inference)

incoming frames. This emulates a realistic behavior during deployment in which all the frames are processed sequentially, yet favoring CoTTA – since a higher framerate, *e.g.* 30FPS, would require CoTTA to adapt on even fewer frames to keep the pace. The video sequence is unlabelled, so we cannot compute mIoU and thus we can appreciate our results only qualitatively, nevertheless, we can provide an overview of the adaptation process operated by HAMLET. Figure 8 sketches the domain sequence, domain shift detections, and relative training intervals, *i.e.* 1 when training is active and 0 otherwise. The average speed theoretically obtained by HAMLET in this sequence is 20.4FPS, even though the input rate was capped at 5.88FPS. It is also interesting to notice how sequential frames and underlying natural domain shifts taking place over the video are making the adaptation task significantly more challenging than the Increasing Storm benchmark proposed in [5], nevertheless, HAMLET manages to identify and activate short training burst in correspondence to the domain shifts, enough to vouch for effective adaptation to the new domains encountered.

In the second video (https://www.youtube.com/watch?v=zjxPbCphPDE) we showcase HAMLET in action in a real environment – i.e., on the road from Seoul to Daegu, Korea. During the trip, we face several different domain transitions, meeting heavy rain, highway environment, dusk, and even nighttime. This latter qualitative result proves that, despite most experiments in the main paper having been conducted in semi-synthetic datasets, HAMLET is effective on real data as well and can be effectively deployed for real applications. The video shows, on top, the input RGB images from the sequence being processed, and at the bottom, the results by SegFormer trained on the source domain (left) and HAMLET being adapted on the sequence itself (right). First and foremost, we point out how the video itself exposes several domain shifts due to the environment itself – i.e., SegFormer has been trained on Cityscapes, featuring cities from Germany in a mostly urban environment, while the whole video features Korea and transits from urban roads to highways. We can appreciate how these domain shifts do not represent a challenge for HAMLET. Then, we observe that rain represents one of the earliest, weather challenges faced in the video, both in the form of small droplets on the glass shield of the car, as well as in actual storms met during driving. While the accuracy of the source SegFormer model dramatically drops in these domains, HAMLET rapidly copes with them and maintains a much higher quality of the results. In the last part of the video, we encounter nighttime domains: despite the much lower brightness in the images and the lower contrast between the different regions (e.g., road vs vegetation or cars), HAMLET can still keep the drop in accuracy moderate, while the source SegFormer model results completely ineffective on such an unseen domain, rarely distinguishing the road from any generic vehicle. Fig. 9 sketches the domain shift detections and relative learning rate schedules, training intervals, *i.e.* 1 when training is active and 0 otherwise. We can observe how HAMLET can identify several domain shifts and tune the adaptation rate accordingly.


}

\end{document}